\documentclass{article}

\PassOptionsToPackage{numbers}{natbib}
\usepackage[preprint]{neurips_2020}
\usepackage{neurips_2020}
\usepackage[utf8]{inputenc} 
\usepackage[T1]{fontenc}    
\usepackage{hyperref}       
\usepackage{url}            
\usepackage{booktabs}       
\usepackage{amsfonts}       
\usepackage{nicefrac}       
\usepackage{microtype}      

\title{Compressed Sensing with Deep Image Prior and Learned Regularization}

\author{
  Dave Van Veen\thanks{Equal contribution} \thanks{Department of Electrical and Computer Engineering, University of Texas. Austin, TX}\\
  \texttt{vanveen@utexas.edu} \\
  \And
  Ajil Jalal\footnotemark[1] \footnotemark[2] \\
  \texttt{ajiljalal@utexas.edu} \\
  \And 
    Mahdi Soltanolkotabi \thanks{Department of Electrical Engineering, University of Southern California. Los Angeles, CA}\\
  \texttt{soltanol@usc.edu} \\ 
  \AND
  Eric Price \thanks{Department of Computer Science, University of Texas. Austin, TX}\\
  \texttt{ecprice@cs.utexas.edu} \\
    \And
  Sriram Vishwanath \footnotemark[2]\\
  \texttt{sriram@austin.utexas.edu} \\
    \And
  Alexandros G. Dimakis \footnotemark[2]\\
  \texttt{dimakis@austin.utexas.edu} \\
}

\usepackage{tikz}
\usepackage{algorithm}
\usepackage{algorithmic}
\usepackage{bm}

\usepackage{amsfonts,amsmath,amsthm,amssymb}
\usepackage{graphicx}
\usepackage{caption}
\usepackage{subcaption}
\usepackage{xcolor}
\usepackage{multirow}

\usepackage{comment}
\usepackage{arydshln}
\usepackage{dblfloatfix}
\usepackage{color,soul}
\DeclareMathOperator*{\argmin}{arg\,min}
\DeclareMathOperator*{\argmax}{arg\,max}

\newcommand{\R}{\mathbb{R}}

\newcommand{\h}[1]{\hat{#1}}

\newtheorem{theorem}{Theorem}[section]

\newtheorem{lemma}[theorem]{Lemma}

\newcommand{\bteta}{\theta}
\newtheorem{assumption}{Assumption}
\newcommand{\E}{\operatorname{\mathbb{E}}}
\newcommand{\twonorm}[1]{\left\|#1\right\|}
\newcommand{\opnorm}[1]{\left\|#1\right\|}
\newcommand{\sgn}[1]{\textrm{sgn}(#1)}

\begin{document}

\maketitle

\begin{abstract}
We propose a novel method for compressed sensing recovery using
untrained deep generative models. Our method is based on the recently
proposed Deep Image Prior (DIP), wherein the convolutional weights of
the network are optimized to match the observed measurements. We show
that this approach can be applied to solve any differentiable linear inverse
problem, outperforming previous unlearned methods. Unlike various learned approaches based on generative models, our method does not require pre-training over large datasets. We further introduce a novel learned regularization technique, which incorporates prior information on the network weights. This reduces reconstruction error, especially for noisy measurements. Finally we prove that, using the DIP optimization approach, moderately overparameterized single-layer networks can perfectly fit any signal despite the nonconvex nature of the fitting problem. This theoretical result provides justification for early stopping.
\end{abstract}
\section{Introduction}
\label{introduction}

We consider the well-studied compressed sensing problem of recovering an unknown signal $x^*\in \R^n$ by observing a set of noisy measurements $y \in \R^m$ of the form
\begin{equation}\label{eqn:y=Ax}
  y=Ax^*+ \eta.
\end{equation}
Here  $A\in \R^{m\times n}$ is a known measurement matrix, typically generated with random independent Gaussian entries. Since the number of measurements $m$ is smaller than the dimension $n$ of the unknown vector $x^*$, this is an under-determined system of noisy linear equations and hence ill-posed. There are many solutions, and some structure must be assumed on $x^*$ to have any hope of recovery. Pioneering research~\citep{donoho2006compressed,candes2006robust,candes2005decoding}
established that if $x^*$ is assumed to be sparse in a known basis, a small number of measurements will be provably sufficient to recover the unknown vector in polynomial time using methods such as Lasso~\citep{tibshirani1996regression}.

Sparsity approaches have proven successful, but more complex models with additional structure have been recently proposed such as model-based compressive sensing~\citep{baraniuk2010model} and manifold models~\citep{hegde2008random,hegde2012signal,eftekhari2015new}.
Bora et al.~\cite{bora2017compressed} showed that deep generative models can be used as excellent priors for images. They also showed that backpropagation can be used to solve the signal recovery problem by performing gradient descent in the generative latent space. This method enabled image generation with significantly fewer measurements compared to Lasso for a given reconstruction error. 
Compressed sensing using deep generative models was further improved in very recent work~\citep{tripathi2018correction,grover2018amortized,kabkab2018task,shah2018solving,fletcher2017inference, DBLP:journals/corr/abs-1802-04073}. Additionally a theoretical analysis of the nonconvex gradient descent algorithm~\citep{bora2017compressed} was proposed by Hand et al.~\cite{hand2017global} under some assumptions on the generative model. 

Inspired by these impressive benefits of deep generative models, we chose to investigate the potential application of such methods for 
medical imaging, a canonical application of compressive sensing. 
A significant problem, however, is that all these previous methods require the existence of \emph{pre-trained} models. While this has been achieved for various types of images, e.g. human faces of CelebA~\citep{liu2015faceattributes} via DCGAN~\citep{radford2015unsupervised}, it remains significantly more challenging for medical images~\citep{wolterink2017generative,schlegl2017unsupervised,nie2017medical, schlemper2017deep}. Instead of addressing this problem in generative models, we found an easier way to circumvent it. 

Surprising recent work by Ulyanov et al.~\cite{ulyanov2017deep} proposed Deep Image Prior (DIP), which uses \emph{untrained} convolutional neural networks. In DIP-based schemes, a convolutional neural network generator (e.g. DCGAN) is initialized with random weights; these weights are subsequently optimized to make the network produce an output as close to the target image as possible. This procedure is unlearned, using no prior information from other images. The prior is enforced only by the fixed convolutional structure of the generator network. 

Generators used for DIP are typically over-parameterized, i.e.~the number of network weights is much larger compared to the output dimension. For this reason DIP 
has empirically been found to overfit to noise if run for too many iterations: The reconstruction error initially decreases and then plateaus, at approximately $500$ iterations, as the network fits the original image. Then, at roughly $10,000$ iterations, the error decreases further, as the network starts fitting the noise~\citep{ulyanov2017deep}. Early stopping is a heuristic intended to terminate the optimization procedure within this plateau region, 
and avoid overfitting to noise. In this paper we theoretically prove that this overfitting phenomenon occurs with gradient descent for any signal and hence justify the use of early stopping and other regularization methods. 

\textbf{Our Contributions:} 
\begin{itemize}
    \item In Section~\ref{methods} we propose DIP for compressed sensing (CS-DIP). Our basic method is as follows. Initialize a DCGAN generator with random weights; use gradient descent to optimize these weights such that the network produces an output which agrees with the observed measurements as much as possible. This unlearned method can be improved with a novel \textit{learned regularization} technique, which regularizes the DCGAN weights throughout the optimization process. 
     \item  
     In Section~\ref{theory} we theoretically prove that DIP will fit any signal to zero error with gradient descent. Our result is established for a network with a single hidden layer and sufficient constant fraction over-parametrization. While it is expected that over-parametrized neural networks can fit any signal, the fact that gradient descent can provably solve this non-convex problem is interesting and provides theoretical justification for early stopping, a phenomenon justified empirically by Ulyanov et al.~\cite{ulyanov2017deep}.
    \item In Section~\ref{experiments_results} we empirically show that CS-DIP outperforms previous unlearned methods in many cases. While pre-trained or ``learned'' methods frequently perform better~\citep{bora2017compressed}, we have the advantage of not requiring a generative model trained over large datasets. As such, we can apply our method to various medical imaging datasets for which data acquisition is expensive and generative models are difficult to train.
\end{itemize}
\section{Background}
\label{background}

\subsection{Compressed Sensing: Classical and Unlearned Approaches}

A classical assumption made in compressed sensing is that the vector $x^*$ is $k$-sparse in some basis such as wavelet or discrete cosine transform (DCT). Finding the sparsest solution to an underdetermined linear system of equations is NP-hard in general; however, if the matrix $A$ satisfies conditions such as the Restricted Eigenvalue Condition (REC) or Restricted Isometry Property (RIP)~\citep{candes2006stable, bickel2009simultaneous, donoho2006compressed, tibshirani1996regression}, then
$x^*$ can be recovered in polynomial time via convex relaxations~\citep{tropp2006just} or iterative methods. There is extensive compressed sensing literature regarding assumptions on $A$, numerous recovery algorithms, and variations of RIP and REC~\citep{bickel2009simultaneous, negahban2009unified, agarwal2010fast, bach2012optimization, loh2011high}.

Compressed sensing methods have found many applications in imaging, for example the single-pixel camera (SPC)~\citep{duarte2008single}. Medical tomographic applications include x-ray radiography, microwave imaging, magnetic resonance imaging (MRI)~\citep{winters2010sparsity, chen2008prior, lustig2007sparse}. Obtaining measurements for medical imaging can be costly, time-consuming, 
and in some cases dangerous to the patient~\citep{qaisar2013compressive}. 
As such, an important goal is to reduce the number of measurements while maintaining good reconstruction quality. 

Aside from the classical use of sparsity, recent work has used other priors to solve linear inverse problems. Plug-and-play priors~\citep{venkatakrishnan2013plug,chan2017plug} and Regularization by Denoising~\citep{romano2017little} have shown how image denoisers can be used to solve general linear inverse problems. A key example of this is BM3D-AMP, which applies a Block-Matching and 3D filtering (BM3D) denoiser
to an Approximate Message Passing (D-AMP) algorithm~\citep{metzler2016denoising,metzler2015bm3d}. AMP has also been applied to linear models in other contexts~\citep{schniter2016vector}. Another related algorithm
is TVAL3~\citep{zhang2013improved, li2009user} which leverages augmented Lagrangian multipliers to achieve impressive performance on compressed sensing problems. 
In many different settings, we compare our algorithm to these prior methods: BM3D-AMP, TVAL3, and Lasso. 

\subsection{Compressed Sensing: Learned Approaches}

While sparsity in some chosen basis is well-established, recent work has shown better empirical performance when neural networks are used \citep{bora2017compressed}. This success is attributed to the fact that neural networks are capable of learning image priors from very large datasets~\citep{goodfellow2014generative, kingma2013auto}. There is significant recent work on solving linear inverse problems using various learned techniques, e.g.~recurrent generative models~\citep{mardani2017recurrent} and auto-regressive models~\citep{dave2018solving}. Additionally approximate message passing (AMP) has been extended to a learned setting by Metzler et al.~\cite{metzler2017learned}.

Bora et al.~\cite{bora2017compressed} is the closest to our set-up. In this work the authors assume that the unknown signal is in the range of a pre-trained generative model such as a generative adversarial network (GAN)~\citep{goodfellow2014generative} or variational autoencoder (VAE)~\citep{kingma2013auto}. 
The recovery of the unknown signal is obtained via gradient descent in the latent space by searching for a signal that satisfies the measurements. This can be directly applied for linear inverse problems and more generally to any differentiable measurement process.  Recent work has built upon these methods using new optimization techniques~\citep{Chang17}, uncertainty autoencoders~\citep{grover2018uncertainty},
and other approaches~\citep{dhar2018modeling, kabkab2018task, mixon2018sunlayer, pandit2019asymptotics, rusu2018meta, hand2018phase}. The key point is that all this prior work requires pre-trained generative models, in contrast to CS-DIP. 
Finally, there is significant ongoing work to understand DIP and develop related approaches~\citep{heckel2018decoder, heckel2018deep, dittmer2018regularization}.
\section{Proposed Algorithm}
\label{methods}

Let $x^*\in\R^n$ be the signal that we are trying to reconstruct, $A\in\R^{m\times n}$ be the measurement matrix, and $\eta\in\R^m$ be independent noise. Given the measurement matrix $A$ and the observations $y=Ax^* + \eta$, we wish to reconstruct an $\h{x}$ that is close to $x^*$.

A generative model is a deterministic function $G(\cdot;w){:} \; \R^k\rightarrow\R^n$ which takes as input $z\in \R^k$ and is parameterized by ``weights'' $w\in\R^d$, producing an output $G(z;w)\in\R^n$. These models have shown excellent performance generating real-life signals such as images~\citep{goodfellow2014generative,kingma2013auto} and audio~\citep{wavenet}. We investigate deep convolutional generative models, a special case in which the model architecture has multiple cascaded layers of convolutional filters~\citep{krizhevsky2012imagenet}. In this paper we 
restrict the signal to be images and apply a DCGAN~\citep{radford2015unsupervised} model. This model architecture contrasts with Ulyanov et al.~\cite{ulyanov2017deep}, who employ a U-net~\cite{ronneberger2015u}. By using a different convolutional architecture, we further support the working hypothesis that network structure, not representation learning, is the key component in image reconstruction. Our choice of DCGAN also allows for a clearer comparison in the CS setting, e.g.~learned methods of Bora et al.~\cite{bora2017compressed} which employ this architecture. Lastly, we've found DCGAN to have $5{\text -}10\times$ faster runtime than U-net.

\subsection{Compressed Sensing with Deep Image Prior (CS-DIP)}
Our approach is to find a set of weights for the convolutional network such that the measurement matrix applied to the network output, i.e. $AG(z;w)$, matches the measurements $y$ we are given. Hence we initialize an \emph{untrained} network $G(z;w)$ with some fixed $z$ and solve the following:
\begin{equation}\label{eqn:w_star}
    w^*=\argmin_w  \| y - AG(z;w) \|^{2}.
\end{equation}

This is, of course, a non-convex problem because $G(z;w)$ is a complex feed-forward neural network. Still we can use  gradient-based optimizers for any generative model and measurement process that is differentiable.  
Generator networks such as DCGAN are biased toward smooth, natural images due to their convolutional structure; thus the network structure alone provides a good prior for reconstructing images in problems such as inpainting and denoising~\citep{ulyanov2017deep}.
Our finding is that this applies to general linear measurement processes.
Furthermore, our method also directly applies to any differentiable forward operator $A$. 
We restrict our solution to lie in the span of a convolutional neural network. If a sufficient number of measurements $m$ is given, we obtain an output such that $x^*\approx G(z;w^*)$. 

Note that this method uses an untrained generative model and optimizes over the network weights $w$. In contrast previous methods, such as that of Bora et al.~\cite{bora2017compressed}, use a trained model and optimize over the latent $z$-space, solving $z^*=\argmin_z  \| y - AG(z;w) \|^{2}$. We instead initialize a random $z$ with Gaussian i.i.d. entries and keep this fixed throughout the optimization process.

In our algorithm we leverage the well-established total variation regularization~\citep{rudin1992nonlinear,wang2008new, liu2018image}, denoted as $TV(G(z;w))$. We also propose an additional learned regularization technique, $LR(w)$; note that without this technique, i.e. when $\lambda_L = 0$, our method is completely unlearned. Lastly we use early stopping, a phenomenon that will be justified theoretically in Section~\ref{theory}.

Thus the final optimization problem becomes 
\begin{align}
\label{eqn:y-AG}
    w^*=   \argmin_w  &  \| y - A\, G(z;w) \|^{2} +  R(w; \lambda_{T}, \lambda_{L}). 
\end{align}
The regularization term contains hyperparameters $\lambda_{T}$ and $\lambda_{L}$
for total variation and learned regularization: 
$R(w; \lambda_{T}, \lambda_{L})= \lambda_{T} TV(G(z;w)) + \lambda_{L} LR(w)$. Next we discuss this $LR(w)$ term.

\subsection{Learned Regularization}\label{sec: learned regularization}

Without learned regularization CS-DIP relies only on linear measurements taken from one unknown image. We now introduce a novel method which leverages a small amount of training data to optimize regularization.
In this case training data refers to measurements from additional ground truth of a similar type, e.g.~measurements from other x-ray images.

To leverage this additional information, we pose Eqn.~\ref{eqn:y-AG} as a Maximum a Posteriori (MAP) estimation problem and propose a novel prior on the weights of the generative model. This prior then acts as a regularization term, penalizing the model toward an optimal set of weights $w^*$.

For a set of weights $w\in\R^d$, we model the \emph{likelihood} of the measurements $y=Ax, y\in\R^m,$ and the prior on the weights $w$ as Gaussian distributions given by
\[
p(y|w) = \frac{\exp\left(-\dfrac{\|y-AG(z;w)\|^2}{2\lambda_L}\right)}{\sqrt{(2\pi \lambda_L)^m}}; \quad \quad \quad
p(w) = \frac{\exp\left(-\frac{1}{2}\left(w-\mu\right)^T\Sigma^{-1}\left(w-\mu\right)\right)}{\sqrt{(2\pi)^d|\Sigma|}},
\]

where $\mu\in\R^d$ and $\Sigma\in\R^{d\times d}$.

In this setting we want to find a set of weights $w^*$ that maximizes the log posterior on $w$ given $y$, i.e.,
\begin{eqnarray}\label{eqn: y-AG + R}
    w^*& = & \argmax_w p(w|y) \equiv  \argmin_w \|y-AG(z;w)\|^2 + \, \lambda_L \left(w-\mu\right)^T\Sigma^{-1}\left(w-\mu\right).
\end{eqnarray}

This gives us the learned regularization term
\begin{equation}\label{eqn: lr}
LR(w) = \left(w-\mu\right)^T\Sigma^{-1}\left(w-\mu\right),    
\end{equation}

where the coefficient $\lambda_L$ in Eqn.~\ref{eqn: y-AG + R} controls the strength of the prior.

Our motivation for assuming a Gaussian distribution on the weights is to build upon the proven success of $\ell_2$ regularization, which also makes this assumption. Notice that when $\mu=0$ and $\Sigma=I_{d\times d},$ this regularization term is equivalent to $\ell_2$-regularization. Thus this method can be thought of as an adaptive version of standard weight decay. Further, because the network weights are initialized Gaussian i.i.d., we assumed the optimized weights would also be Gaussian. Previous work has shown evidence that the convolutional weights in a trained network do indeed follow a Gaussian distribution~\citep{ma2018invertibility}.

\subsubsection{Learning the Prior Parameters}\label{sec:learning_mu_sig}
In the previous section, we introduced the learned regularization term $LR(w)$ defined in Eqn.~\ref{eqn: lr}. 
However we have not yet learned values for parameters $(\mu,\Sigma)$ that incorporate prior knowledge of the network weights. We now propose a way to estimate these parameters.

Assume we have a set of measurements $S_Y = \{y_1,y_2,\cdots,y_Q\}$ from $Q$ different images $S_X = \{x_1,x_2,\cdots,x_Q\}$, each obtained with a different measurement matrix $A$. For each measurement $y_q,q\in \{1,2,...,Q\},$ we run CS-DIP to solve the optimization problem in Eqn.~\ref{eqn:y-AG} and obtain an optimal set of weights $W^*=\{w^*_1,w^*_2,\cdots,w^*_Q\}$. Note that when optimizing for the weights $W^*,$ we only have access to the measurements $S_Y$, not the ground truth $S_X$.

The number of weights $d$ in deep networks tends to be very large. As such, learning a distribution over each weight, i.e. estimating $\mu \in \R^d $ and $\Sigma \in \R^{d\times d}$, becomes intractable. We instead use a layer-wise approach: with $L$ network layers, we have $\mu \in \R^L $ and $\Sigma \in \R^{L\times L}$. Thus each weight within layer $l\in \{1,2,...,L\}$ is modeled according to the same $\mathcal{N}(\mu_l, \, \Sigma_{ll}) $ distribution. For simplicity we assume $\Sigma_{ij} = 0 \; \forall \, i \neq j$, i.e. that network weights are independent across layers. The process of estimating statistics $(\mu,\Sigma)$ from $W^*$ is described in Algorithm~\ref{algo:DIP-musig}. 

We use this learned $(\mu,\Sigma)$ in the regularization term $LR(w)$ from Eqn.~\ref{eqn: lr} for reconstructing measurements of images. We refer to this technique as \emph{learned regularization}. While this may seem analogous to batch normalization~\citep{ioffe2015batch}, note that we only use $(\mu,\Sigma)$ to penalize the $\ell_2$-norm of the weights and do not normalize the layer outputs themselves.

\subsubsection{Discussion of Learned Regularization}

The proposed CS-DIP does not require training if no learned regularization is used, i.e. if $\lambda_L = 0$ in Eqn.~\ref{eqn:y-AG}. 
This means that CS-DIP can be applied only with measurements from a single image and no prior information of similar images in a dataset. 

Our next idea, learned regularization, utilizes a small amount of prior information,  requiring access to measurements from a small number of similar images (roughly $10$). In contrast, other pre-trained models such as that of Bora et al.~\cite{bora2017compressed} require access to ground truth from a massive number of 
similar images (tens of thousands for CelebA). If such a large dataset is available, and if a good generative model can be trained on that dataset, we expect that pre-trained models
would outperform our method. Our approach is instead more suitable for reconstructing problems where large amounts of data or good generative models are not readily available. 
\section{Theoretical Results}
\label{theory}

In this section we provide theoretical evidence to highlight the importance of early stopping for DIP-based approaches. Here we focus on denoising a noisy signal $y\in\R^m$ by optimizing over network weights. This problem takes the form:
\begin{equation}
\min_w\text{ }\mathcal{L}(w):=\| y - AG(z;w) \|^{2}.
\end{equation}
We focus on generators 
consisting of a single hidden-layer ReLU network with $k$ inputs, $d$ hidden units, and $n$ outputs. Using $w=(W,V)$ the generator model in this case is given by
\begin{align}
\label{genmodel}
 G(z;W,V)=V\cdot\text{ReLU}(Wz),  
\end{align}
where $z\in\R^k$ is the input, $W\in\R^{d\times k}$ the input-to-hidden weights, and $V\in\R^{n\times d}$ the hidden-to-output weights. We hence train over $W$ using gradient descent and assume $V$ is fixed\footnote{We note that our theoretical framework can also allow for training over $V$. In fact, since the problem is quadratic over $V$, this analysis is in principle not much more difficult. However, we avoid this as it makes our proofs more difficult to follow without adding any further insight.}.
With these formulations in place, we are now ready to state our theoretical result.
\begin{theorem}\label{mainthm}
Consider fitting measurements $AG(z;W,V)$ from the output of a generator of the form $W\mapsto G(z;W,V)=V\cdot ReLU\left(Wz\right)$ to a signal $y\in\R^m$ with $A\in\R^{m\times n}$, $z\in\R^k$, $W\in\R^{d\times k}$, $V\in\R^{n\times d}$, and $ReLU(z)=\max(0,z)$. Furthermore, let $A$ be a matrix with orthonormal rows (i.e.~$AA^T=I_m$)
and assume $V$ is a random matrix with i.i.d.~$\mathcal{N}(0,\nu^2)$ entries with $\nu=\frac{1}{\sqrt{dm}}\frac{\twonorm{y}}{\twonorm{z}}$. Starting from an
initial weight matrix $W_0$ selected at random with i.i.d.~$\mathcal{N}(0, 1)$ entries, we run gradient descent updates of the form $W_{\tau+1}=W_\tau-\eta \nabla \mathcal{L}(W_\tau)$ on the loss 
\begin{align*}
\mathcal{L}(W)=\frac{1}{2}\twonorm{AV\cdot\text{ReLU}\left(Wz\right)-y}^2,
\end{align*}
with step size $\eta=\frac{\bar{\eta}}{\twonorm{y}^2}\frac{8m}{4m+d}$ where $\bar{\eta}\le 1$. Assuming that $d\ge Cm,$
with $C$ a fixed numerical constant, then 
\begin{align*}
\twonorm{AV\cdot\text{ReLU}\left(W_\tau z\right)-y}\leq 3\left(1-\frac{\bar{\eta}}{8(4m+d)}\right)^{\tau}\twonorm{y}
\end{align*}
holds for all $\tau$ with probability at least $1-5e^{-m/2}-e^{-d/2}-e^{-4d^{\frac{2}{3}}m^{\frac{1}{3}}}$.
\end{theorem}
Our theoretical result shows that after many iterative updates,
gradient descent will solve this non-convex optimization problem and fit any signal $y$, if the generator network is sufficiently wide. This occurs as soon as the number of hidden units $d$ exceeds the signal size $n$ by a constant factor. 
Our theorem directly applies to many compressed sensing measurement matrices, in particular any matrix obtained by subsampling the rows of an orthonormal matrix (e.g. sub-sampling a Fourier matrix). This is possible because, for any such orthonormal matrix, $A V$ has the same distribution as a Gaussian matrix with i.i.d. entries. 
This result demonstrates that early stopping is necessary for DIP-based methods to be successful; otherwise the network can fit any signal, including one that is noisy.

Our proof builds on theoretical ideas from Oymak et al.~\cite{oymak2019towards} which provide a general framework for establishing global convergence guarantees for overparameterized nonlinear learning problems based on various properties of the Jacobian mapping along the gradient descent trajectory. While our proof leverages relevant prior work~\cite{du2018gradient, Oymak:2018aa}, our argument is quite specialized and intricate with new techniques such as Gordon's Lemma. This allows us to have moderate network overparameterization that is only linear in the number of measurements, contrary to other results in the literature which require a significant amount of overparameterization. Ultimately we combine  tools from empirical process theory, random matrix theory, and matrix algebra to show that, starting from a random initialization, the Jacobian mapping across all iterates has favorable properties with high probability, hence facilitating convergence to a global optima.
\section{Experiments}
\label{experiments_results}

\subsection{Experimental Setup}\label{expt:gaussian}

\textbf{Measurements:} We evaluate our algorithm using two different measurements processes, i.e. matrices $A\in \R^{m\times n}$. First we set the entries of $A$ to be Gaussian i.i.d. such that $A_{i,j} \sim \mathcal{N}(0, \, \frac{1}{m})$. Recall $m$ is the number of measurements, and $n$ is the number of pixels in the ground truth image. This measurement process is standard practice in compressed sensing literature; hence we use it on each dataset. Additionally we use a Fourier measurement process common in MRI applications~\citep{mardani2018neural, mardani2017deep, hammernik2018learning, lehtinen2018noise2noise, lustig2008compressed}.

\textbf{Datasets:} We use our algorithm to reconstruct both grayscale and RGB images. For grayscale we use the first 100 images in the test set of MNIST~\citep{lecun1998gradient} and also 60 random images from the Shenzhen Chest X-Ray Dataset~\citep{jaeger2014two}, downsampling a $512 \times 512$ crop to $256 \times 256$ pixels. For RGB we use retinopathy images from the STARE dataset~\citep{hoover2000locating} with $512 \times 512$ crops downsized to $128 \times 128$ pixels.

\textbf{Baselines:} We compare our algorithm to state-of-the-art unlearned methods such as BM3D-AMP~\citep{metzler2016denoising, metzler2015bm3d}, TVAL3~\citep{li2011compressive, li2009user, zhang2013improved}, and Lasso in a DCT basis~\citep{ahmed1974discrete}. We also evaluated the performance of Lasso in a Daubechies wavelet basis~\citep{daubechies1988orthonormal, wasilewski2010pywavelets} but found this performed worse than Lasso - DCT on all datasets. Thus hereon we refer to Lasso - DCT as ``Lasso'' and do not include results of Lasso - Wavelet. We used sci-kit learn~\citep{scikit-learn} for the implementation of Lasso and code provided by the original authors for BM3D-AMP and TVAL3. A standard grid search was performed over each baseline to tune hyperparameters.

\textbf{Metrics:} To quantitatively evaluate the performance of our algorithm, we use per-pixel mean-squared error (MSE) between the reconstruction $\hat{x}$ and true image $x^*$, i.e. $\frac{\| \hat{x} - x^* \|^2}{n}$. Note that because these pixels are over the range $[-1,1]$, it's possible for MSE to be greater than $1$.

\textbf{Implementation:}
To find a set of weights $w^*$ that minimize Eqn.~\ref{eqn:y-AG}, we use PyTorch~\citep{paszke2017automatic} with a DCGAN architecture. Our network has depth 7 and uses convolutional layers with ReLU activations. We use the RMSProp optimizer~\citep{tieleman2012lecture} with learning rate $10^{-3}$, momentum $0.9$, and $1000$ update steps for every set of measurements. These parameters are the same across all datasets. We initialize one random measurement matrix $A$ for each image.

\begin{table}[!t]
    \caption{Evaluating the benefits of learned regularization (LR) on x-ray images with varying levels of noise and number of measurements. Table values are percent decrease in error, e.g. at  $\sigma^2_{\eta} = 0$ and $m=500$, LR reduces MSE by $9.9\%$. The term $\sigma^2_{\eta}$ corresponds to variance of the noise vector $\eta$ in Eqn.~\ref{eqn:y=Ax}, i.e.~each entry of $\eta$ is drawn independently $\mathcal{N}(0, \, \nicefrac{\sigma^2_{\eta}}{m})$. These results indicate that LR tends to provide greater benefit with noisy signals and with fewer measurements.
    }
    \label{table:lreg_mse}
  \centering
  \begin{tabular}{llllll}
    \toprule
    \multicolumn{4}{r}{Measurements, $m$}                   \\
    \cmidrule(r){2-6}
    $\sigma^2_{\eta}$ & 500 & 1000 & 2000 & 4000 & 8000 \\    
    \midrule
    0  & 9.9\% & 2.9\% & 0.2\% & 2.0\% & 0.6\% \\
    10  & 11.6\% & 4.6\% & 4.5\% & 2.4\% & 1.0\% \\
    100  & 14.9\% & 19.2\% & 5.0\% & 3.9\% & 2.8\% \\
    1000  & 37.4\% & 30.6\% & 19.8\% & 3.0\% & 6.2\% \\
    \bottomrule
  \end{tabular}
\end{table}

More implementation details can be found in the appendix, such as hyperparameter search, network initializations, and early stopping criterion. Code for these experiments is available in our GitHub repository: \url{github.com/davevanveen/compsensing_dip}.

\subsection{Experimental Results}

\subsubsection{Results: Learned Regularization}\label{sec:lr_results}

We first evaluate the benefits of learned regularization by comparing our algorithm with and without learned regularization, i.e. $\lambda_L = 100$ and $\lambda_L = 0$, respectively, while all other parameters across this comparison are held constant. The latter setting of $\lambda_L = 0$ is an unlearned method, as we are not leveraging ($\mu, \Sigma$) from a specific dataset. In the former setting of $\lambda_L = 100$, we first learn ($\mu, \Sigma$) from a particular set of ten x-ray images; we then evaluate on a different set of x-ray images. We compare these two settings with varying noise and different number of measurements.

Our results in Table~\ref{table:lreg_mse} show that learned regularization does indeed provide benefit, particularly with increasing noise or fewer measurements. Thus we can infer that assuming a learned Gaussian distribution over weights is useful, especially when the original signal is noisy or significantly compressed. In contrast we saw no improvement with vanilla $\ell_2$-regularization, indicating that the benefits of learned regularization can be attributed to prior information of similar images.

\newsavebox\myboxone
\savebox{\myboxone}{\includegraphics[width=0.46\textwidth]{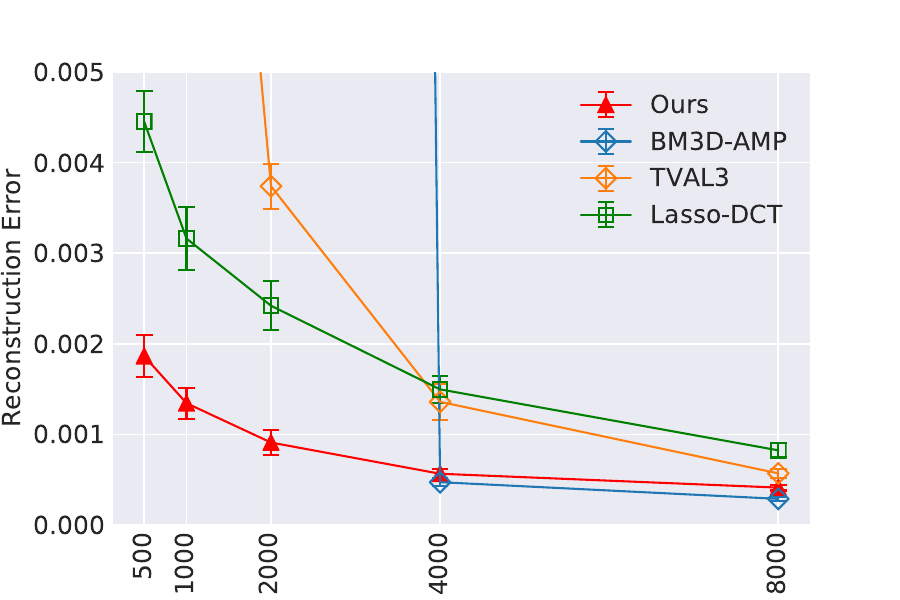}}
\begin{figure*}[t!]
    \begin{subfigure}[b]{0.46\textwidth} 
        \usebox{\myboxone}
        \vspace*{-3mm}
        \caption{MSE - chest x-ray (65536 pixels)}
    \label{fig:graph_x}
    \end{subfigure}\hfill%
    \begin{subfigure}[b]{0.46\textwidth}
        \vbox to \ht\myboxone{%
            \vfill \includegraphics[width=\textwidth]{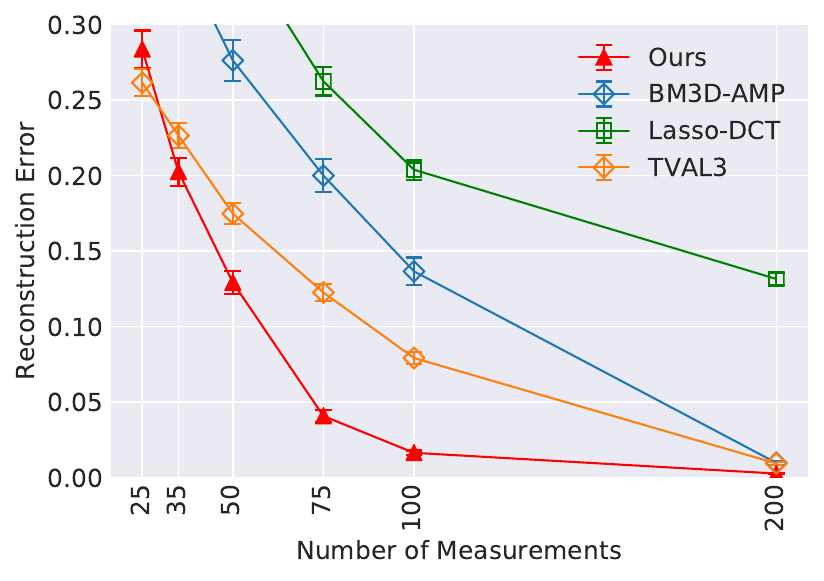}}
        \vspace*{2mm}
        \caption{MSE - MNIST (784 pixels)}
    \label{fig:graph_m}
    \end{subfigure}
    \caption{Per-pixel reconstruction error (MSE) vs.~number of measurements.~Vertical bars indicate 95\% confidence intervals.
    BM3D-AMP frequently fails to converge for fewer than $4000$ measurements on x-ray images, as denoted by error values far above the vertical axis.
    } 
\label{fig:graph_x_m}
\end{figure*}
\newsavebox\myboxtwo
\savebox{\myboxtwo}{\includegraphics[width=0.46\textwidth]{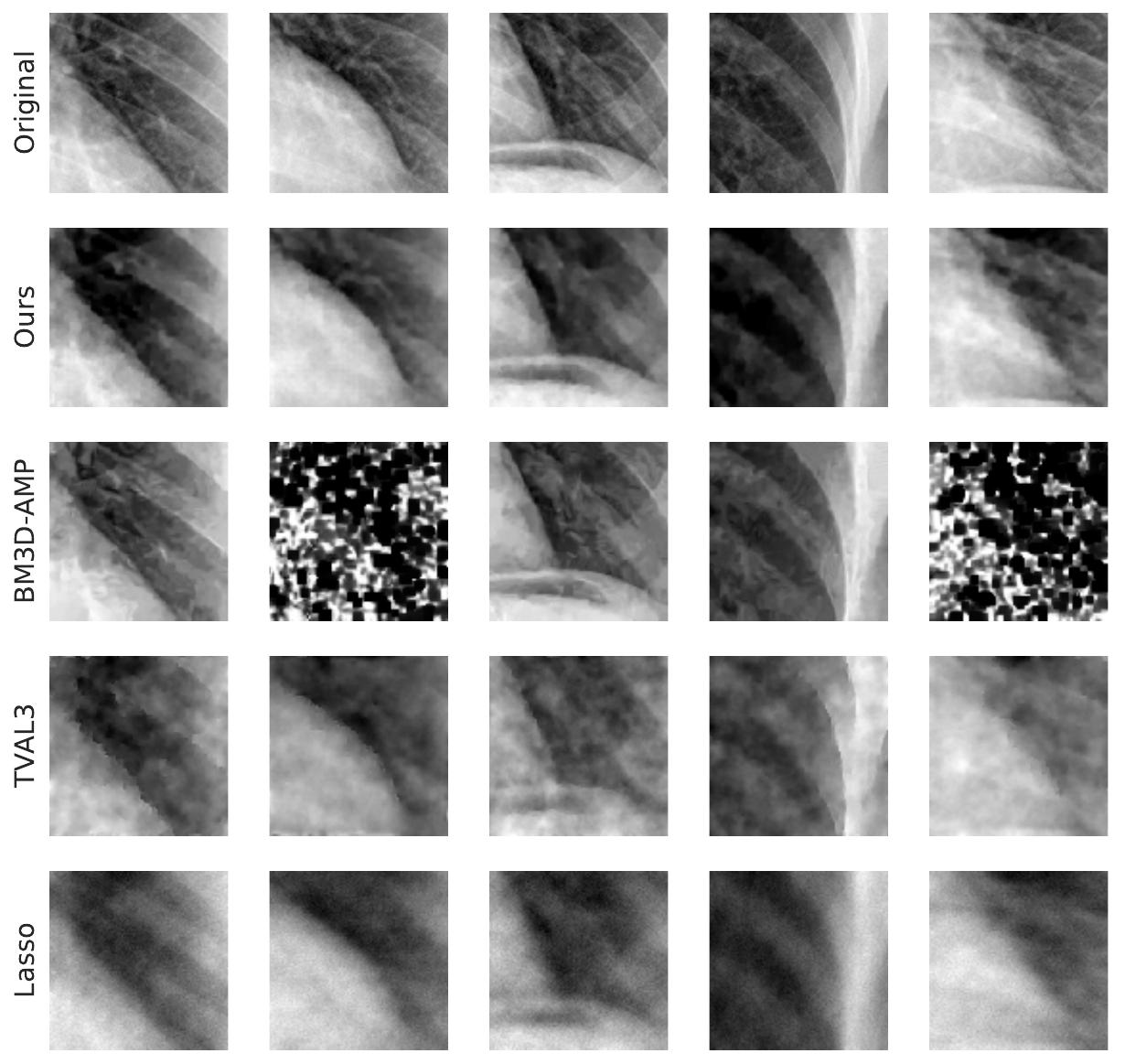}}
\begin{figure*}[b!]
    \begin{subfigure}[t]{0.46\textwidth}
        \usebox{\myboxtwo}  
        \vspace*{-3mm}
        \caption{Reconstructions - chest x-ray}
    \label{fig:recons_x}
    \end{subfigure}\hfill%
    \begin{subfigure}[t]{0.46\textwidth}
        \vbox to \ht\myboxtwo{%
            \vfill \includegraphics[width=\textwidth]{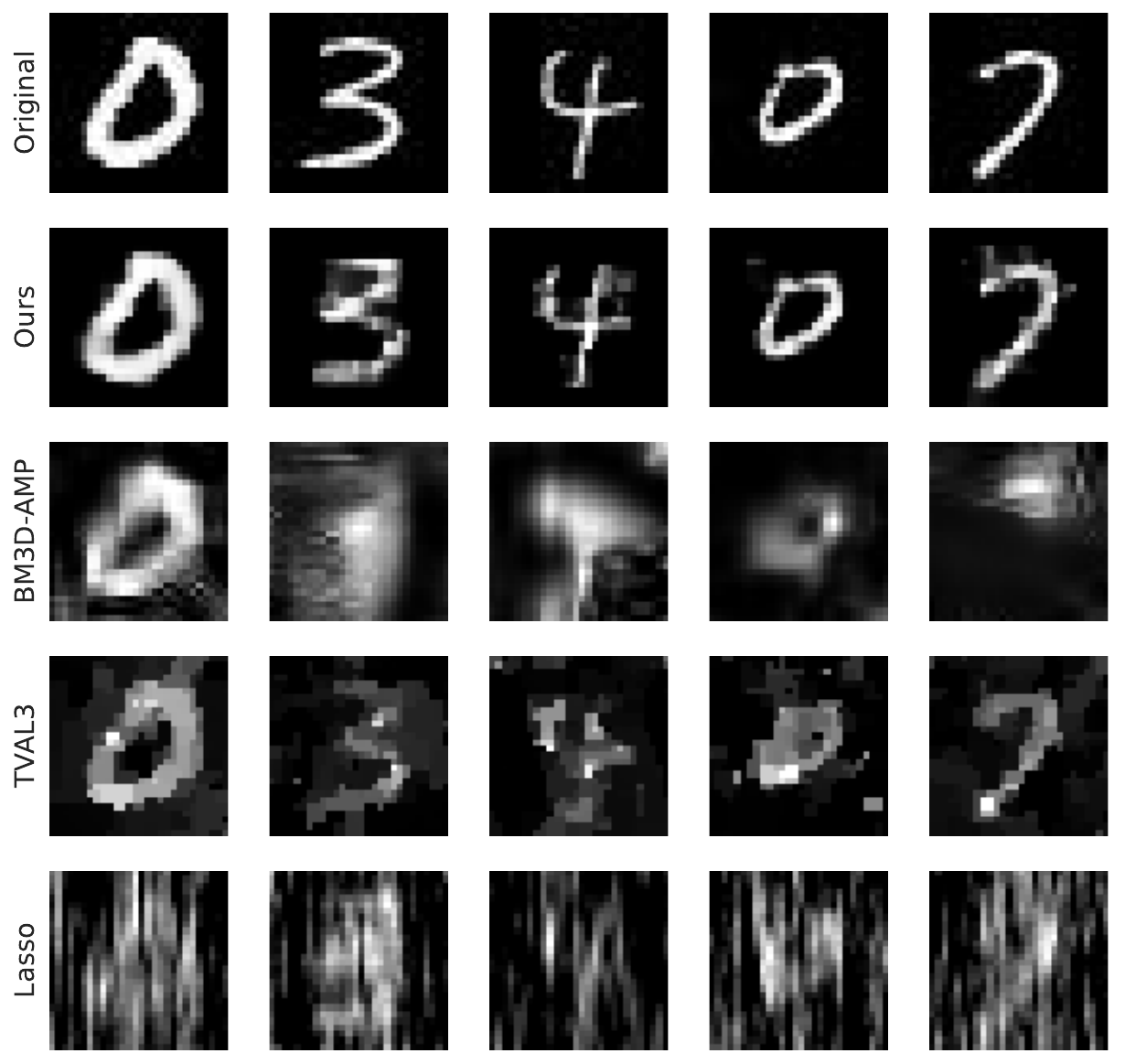}}
        \vspace*{2mm}
        \caption{Reconstructions - MNIST}
    \label{fig:recons_m}
    \end{subfigure}
    \caption{Reconstruction results on x-ray images for m = 2000 measurements (of n = 65536 pixels) and MNIST for m = 75 measurements (of n = 784 pixels). From top to bottom row: original image, reconstructions by our algorithm, then reconstructions by baselines BM3D-AMP, TVAL3, and Lasso. For x-ray images the number of measurements obtained are 3\% the number of pixels (i.e. $\frac{m}{n} = .03$), for which BM3D-AMP often fails to converge.
    }
\label{fig:recons_x_m}
\end{figure*}

\subsubsection{Results: Unlearned CS-DIP}
For the remainder of this section, we evaluate our algorithm in the noiseless case without learned regularization, i.e.~when $\eta = 0$ in Eqn.~\ref{eqn:y=Ax} and $\lambda_L = 0$ in Eqn.~\ref{eqn:y-AG}. Hence CS-DIP is completely unlearned; as such, we compare it to other state-of-the-art unlearned algorithms on various datasets and with different measurement matrices.

\textbf{MNIST:}
\label{sec:mnist_results} In Figure~\ref{fig:graph_m} we plot reconstruction error with varying number of measurements $m$ of $n$ = 784. This demonstrates that our algorithm outperforms baselines in almost all cases. Figure~\ref{fig:recons_m} shows reconstructions for 75 measurements, while remaining reconstructions are in the appendix.

\textbf{Chest x-rays:}
\label{sec:xray_results}
In Figure~\ref{fig:graph_x} we plot reconstruction error with varying number of measurements $m$ of $n$ = 65536. Figure~\ref{fig:recons_x} shows reconstructions for 2000 measurements; remaining reconstructions are in the appendix. On this dataset we outperform all baselines except BM3D-AMP for higher $m$. However for lower $m$, e.g.~when the ratio $\frac{m}{n} \leq 3\%$, BM3D-AMP often doesn't converge. This finding seems to support the work of Metzler et al.~\cite{metzler2015bm3d}: BM3D-AMP performs well on higher $m$, e.g. $\frac{m}{n} \geq 10\%$, but recovery at lower sampling rates is not demonstrated. 

\textbf{Comparison to pre-trained DCGAN:} The approach of Bora et al.~\cite{bora2017compressed} similarly employs a DCGAN but is pre-trained over a large dataset. As expected, this method outperforms ours for lower number of measurements $m$; however, as $m$ increases, the pre-trained method's performance saturates while our algorithm continues to improve and consequently outperform its pre-trained counterpart, per Figure~\ref{fig:fig_csgm} in the appendix. This can be attributed to our method optimizing over the weights $w$ as opposed to the latent space $z$, allowing for a more expressive network capable of reconstructing complicated signals e.g.~medical images. The method of Bora et al. is comparatively less expressive; indeed it is only capable of reconstructing simple images, such as MNIST or aligned CelebA~\cite{bora2017compressed}.

\textbf{Additional experiments:}
In the appendix we further demonstrate our algorithm (1) using a Fourier measurement process for $A$ instead of a Gaussian i.i.d. matrix, (2) on RGB retinopathy images, and (3) in the presence of additive noise. We also perform a runtime analysis.

\section{Conclusion}
\label{conclusion}
We demonstrate how Deep Image Prior (DIP) can be generalized to solve any differentiable linear inverse problem, in many cases outperforming state-of-the-art unlearned methods. We further propose learned regularization which 
enforces a learned Gaussian prior on the network weights. This prior reduces reconstruction error, particularly for noisy or compressed measurements.
Lastly we prove that the DIP optimization technique can fit any signal given a sufficiently wide single-layer network. This provides theoretical justification for regularization methods such as early stopping.

\clearpage
\small{\bibliography{main}}
\bibliographystyle{plain}

\clearpage
\normalsize{\newpage
\appendix
\section{Implementation Details}
\label{sec:expmt_details_appendix}

\textbf{Hyperparameter search:} After a standard grid search procedure, we set the TV hyperparameter $\lambda_T = 0.01$, which aids in providing a sharper reconstruction of the image's high frequency components. For the learned regularization experiments in Section~\ref{sec:lr_results}, a similar grid search was performed to set $\lambda_L = 100$. The criteria for selecting a hyperparameter is one that provides lowest error with the observed measurements, i.e. without observing ground truth. We tune hyperparameters on a random set of ten images; this set is disjoint from the images used for evaluation.

\textbf{Initalizations:} The measurement matrix $A$ is initialized at random for each sample. Similarly network input $z$ in Eqn.~\ref{eqn:y-AG} is initialized with random Gaussian i.i.d. entries and then held fixed as we optimize over network weights $w$. We set the dimension of $z$ to be $128$, a standard choice for DCGAN architectures. For a sufficient number of pixels $n$, i.e. for chest x-ray and retinopathy images, different initializations of $z$ do not affect performance. However for smaller $n$, i.e. for MNIST images, performance can vary with different initializations of $z$.

\textbf{Early stopping:} We stop after $1000$ iterations in all experiments. Similar to Figure 2 in Ulyanov et al.~\cite{ulyanov2017deep}, we found MSE to decrease initially and then plateau until roughly $10,000$ iterations, at which point the network overfits to noise. Hence we terminate the optimization procedure within this plateau region to avoid overfitting. This early stopping technique is common in DIP methods, hence our motivation to justify it theoretically in Section~\ref{theory}.

\section{Additional Experiments}

\textbf{Fourier measurement process:}
\label{sec:fourier_results}
All other experiments used a measurement matrix $A$ containing Gaussian i.i.d. entries. We now consider the case where the measurement matrix is a subsampled Fourier matrix. For a 2D image $x$ and a set of indices $\Omega$, the measurements we receive are given by $y_{(i,j)} = [\mathcal{F}(x)]_{(i,j)}, (i,j)\in\Omega$, where $\mathcal{F}$ is the 2D Fourier transform. We choose $\Omega$ to be indices along radial lines, as shown in Figure~\ref{fig: fourier sampling pattern} of the appendix; this choice of $\Omega$ is common in literature~\citep{candes2006robust} and MRI applications~\citep{mardani2017deep,lustig2008compressed, eksioglu2018denoising}. While Fourier subsampling is common in MRI applications, we use it here on images of x-rays simply to demonstrate that our algorithm performs well with different measurement processes.

In Figure~\ref{fig:fourier_graph_nexttor}, we compare our algorithm to baselines on the x-ray dataset for $\{3,5,10,20\}$ radial lines in the Fourier domain, which corresponds to $\{381,634,1260,2500\}$ Fourier coefficients, respectively. Quantitatively we outperform all baselines. Qualitative reconstructions can be found in Figure~\ref{fig: fourier reconstruction 10}.

\textbf{Retinopathy:}
\label{sec:retino_results}
We plot reconstruction error with varying number of measurements $m$ of $n$ = 49152 in Figure~\ref{fig:graph_r_nexttofourier}. On this RGB dataset we quantitatively outperform all baselines except BM3D-AMP on higher $m$; however, even at these higher $m$, patches of green and purple pixels corrupt the image reconstructions as seen in Figure~\ref{fig:recons_r_standalone}. Similar to x-ray for lower $m$, BM3D-AMP often fails to produce anything sensible. All retinopathy reconstructions are located in the appendix.

\textbf{Robustness to noise:} In Figure~\ref{fig:fig_noise} we demonstrate that our algorithm is robust to additive noise, i.e. when $\eta \neq 0$ in Eqn.~\ref{eqn:y=Ax}, achieving similar behavior to baselines.

\textbf{Runtime:} We demonstrate runtimes for all algorithms on the x-ray dataset in Table~\ref{table:runtime-table}. While our algorithm is faster in most cases, we acknowledge this is not a fair comparison as baselines do not have the benefit of running GPU. Meanwhile our algorithm was run on a NVIDIA GTX 1080-Ti. Ultimately this demonstrates that our algorithm executes in a reasonable amount of time, which can be an issue with DIP methods employing a U-net architecture.

\begin{table}
  \caption{Runtime (seconds) for each algorithm with varying number of measurements.}
  \label{table:runtime-table}
  \centering
  \begin{tabular}{lllll}
    \toprule
    Algorithm & 1000 & 2000 & 4000 & 8000 \\    
    \midrule
    CS-DIP    & 15.6 & \textbf{17.1} & \textbf{20.4} & \textbf{29.9} \\
    BM3D-AMP & 51.1 & 54.0 & 67.8 & 71.2 \\
    TVAL3    & \textbf{13.8} & 22.1 & 31.9 & 56.7 \\
    Lasso DCT    & 27.1 & 33.0 & 52.2 & 96.4 \\
    \bottomrule
  \end{tabular}
\end{table}
\newsavebox\myboxthree
\savebox{\myboxthree}{\includegraphics[width=0.46\textwidth]{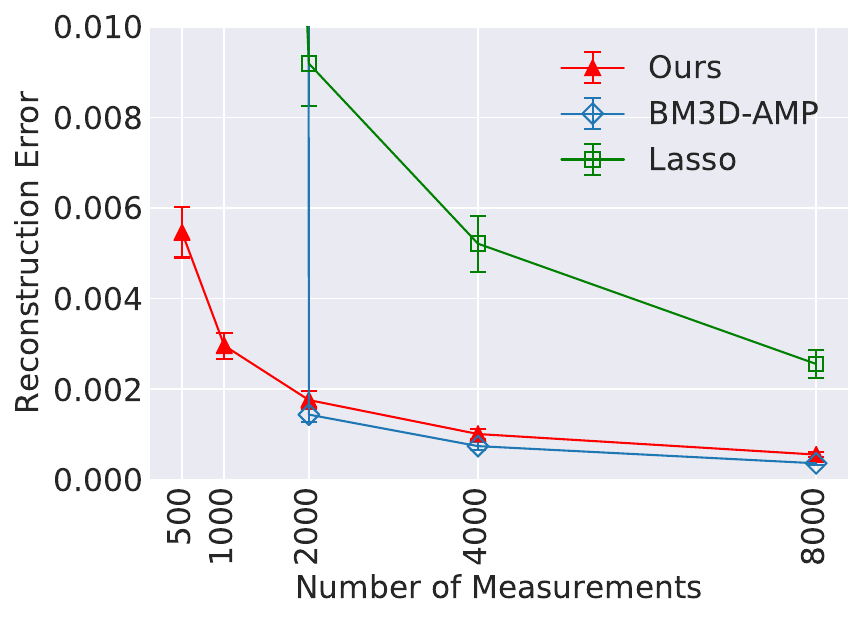}}
\begin{figure*}[b!]
    \begin{subfigure}[t]{0.46\textwidth}
        \usebox{\myboxthree}
        \vspace*{-3mm}
        \caption{MSE - retinopathy (RGB) with Gaussian measurements}
    \label{fig:graph_r_nexttofourier}
    \end{subfigure}\hfill%
    \begin{subfigure}[t]{0.46\textwidth}
        \vbox to \ht\myboxthree{%
            \vfill \includegraphics[width=\textwidth]{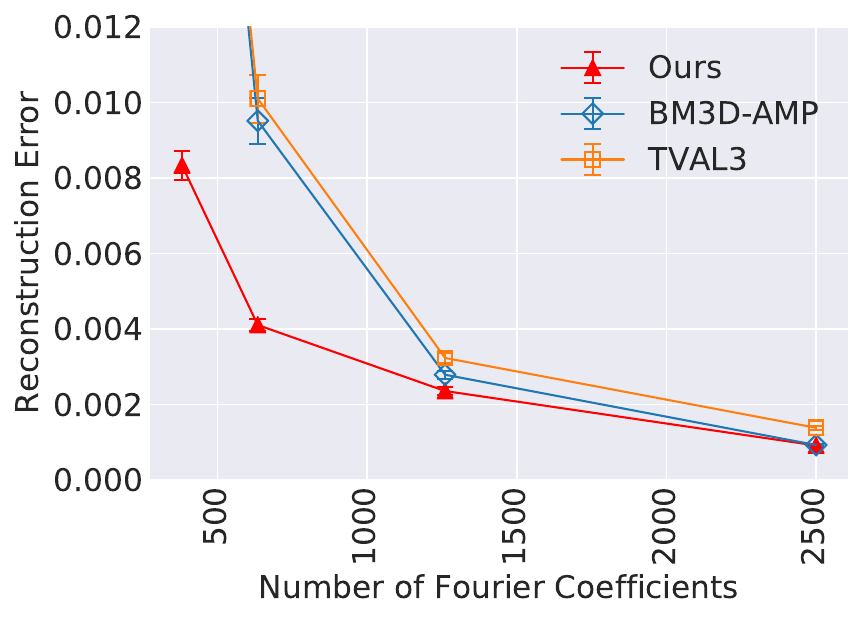}}
        \vspace*{2mm}
        \caption{MSE - chest x-ray with Fourier measurements}
    \label{fig:fourier_graph_nexttor}
    \end{subfigure}
    \caption{Per-pixel reconstruction error (MSE) vs. number of measurements. Vertical bars indicate 95\% confidence intervals. Unfortunately an RGB version of TVAL3 does not currently exist, although related TV algorithms such as FTVd perform similar denoising tasks~\citep{wang2008new}.
    } 
\label{fig:graph_r_f}
\end{figure*}
\newsavebox\myboxnine
\savebox{\myboxnine}{\includegraphics[width=0.46\textwidth]{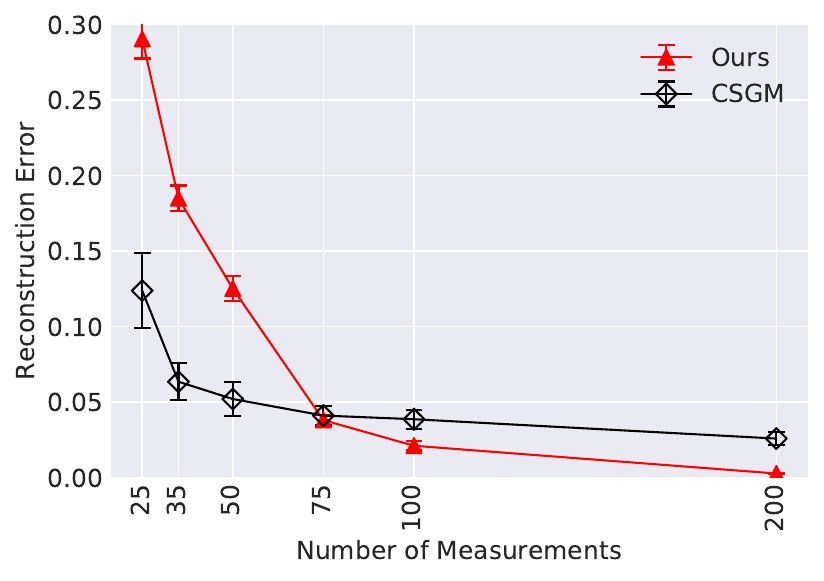}}

\begin{figure*}[b!]
    \begin{subfigure}[t]{0.46\textwidth}
        \usebox{\myboxnine}  
        \caption{Reconstruction error (MSE) on MNIST for varying number of measurements. As expected, the trained algorithm of Bora et al. (CSGM) outperforms our method for fewer measurements; however, CSGM saturates after 75 measurements, as its output is constrained to the range of the generator. 
        This saturation is discussed in Bora et al., Section 6.1.1.
        }
    \label{fig:fig_csgm}
    \end{subfigure}\hfill
    \begin{subfigure}[t]{0.46\textwidth}
        \vbox to \ht\myboxnine{%
            \vfill \includegraphics[width=\textwidth]{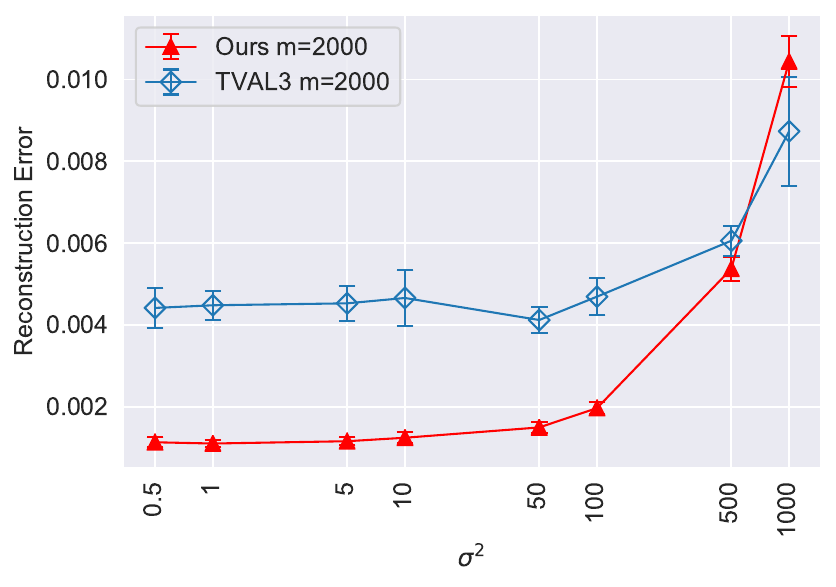}}
        \caption{Reconstruction error (MSE) on x-ray images for varying amounts of noise; number of measurements $m$ fixed at $2000$. The term $\sigma^2$ corresponds to variance of the noise vector $\eta$ in $y = Ax + \eta$, i.e.~each entry of $\eta$ is drawn independently $\mathcal{N}(0, \, \frac{\sigma^2}{m})$. Other baselines have error far above the vertical axis and are thus not visible in this plot.}
    \label{fig:fig_noise}
    \end{subfigure}
\caption{}
\end{figure*}
\begin{algorithm}[t!]
    \caption{Estimate $(\mu$, $\Sigma)$ for a distribution of optimal network weights $W^*$}
    \label{algo:DIP-musig}
    {
    \textbf{Input:} Set of optimal weights $W^*=\{w^*_1,w^*_2,\cdots,w^*_Q\}$ obtained from $L$-layer DCGAN run over $Q$ images; 
    number of samples $S$; number of iterations $T$. \newline
    \textbf{Output:} Mean vector $\mu\in\R^L$; covariance matrix $\Sigma\in \R^{L\times L}$. \newline
    \begin{algorithmic}[1]
    \FOR {$t = 1 \, \text{to} \, T$}
    \STATE Sample $q$ uniformly from $\{1, ..., Q\}$
    \FOR{$l = 1 \, \text{to} \, L$ \COMMENT{for each layer}}
    
    \STATE Get $v \in \R^S$, a vector of $S$ uniformly sampled weights from the $l^{th}$ layer of $w_{q}^*$
    \STATE $M_t[l,:] \leftarrow v^T$ where $M_t[l,:]$ is the $l^{th}$ row of matrix $M_t \in \R^{L\times S}$
    \STATE $\mu_t[l] \leftarrow \frac{1}{S}\sum_{i=1}^{S} v_i $
    \ENDFOR
    \STATE $\Sigma_t \leftarrow \frac{1}{S} M_t M_t^T - \mu_t \mu_t^T$
    \ENDFOR
    \STATE $\mu \leftarrow \frac{1}{T}\sum_{t=1}^{T} \mu_t$
    \STATE $\Sigma \leftarrow \frac{1}{T}\sum_{t=1}^{T} \Sigma_t$
    
    \end{algorithmic}
    }
\end{algorithm}

\clearpage
\newpage
\section{Proof of Section~\ref{theory}: Theoretical Justification for Early Stopping}
In this section we prove our theoretical result in Theorem \ref{mainthm}. We begin with a summary of some notations we use throughout in Section \ref{notation}. Next, we state some preliminary calculations in Section \ref{prelim}. Then, we state a few key lemmas in Section \ref{keyl} with the proofs deferred to Appendix \ref{appproof}. Finally, we complete the proof of Theorem \ref{mainthm} in Section \ref{pfmain}. We note that when $V$ has i.i.d.~Gaussian entries and $A$ contains orthonormal rows, $AV$ also has i.i.d.~Gaussian entries. Therefore without loss of generality we carry out the proof with $A=I$ and $m=n$. The result stated in the theorem simply follows by replacing $V$ in our proof with $AV$.
\subsection{Notation}
\label{notation}
In this section we gather some notation used throughout the proofs. We use $\phi(z)=$ReLU$(z)=\max(0,z)$ with $\phi'(z)=\mathbb{I}_{\{z\ge 0\}}$. For two matrices/vectors $x$ and $y$ of the same size we use $x\odot y$ to denote the entrywise Hadamard product of these two matrices/vectors. We also use $x\otimes y$ to denote their Kronecker product. For two matrices $B\in\R^{n\times d_1}$ and $C\in\R^{n\times d_2}$, we use the Khatrio-Rao product as the matrix $A=B*C\in\R^{n\times d_1d_2}$ with rows $A_i$ given by $A_i=B_i\otimes C_i$. For a matrix $M\in\R^{m\times n}$ we use vect$(M)\in\R^{mn}$ to denote a vector obtained by aggregating the rows of the matrix $M$ into a vector, i.e.~vect$(M)=\begin{bmatrix}M_1 & M_2 & \ldots & M_m\end{bmatrix}^T$. For a matrix $X$ we use $\sigma_{\min}(X)$ and $\|X\|$ denotes the minimum singular value and spectral norm of $X$. Similarly, for a symetric matrix $M$ we use $\lambda_{\min}(M)$ to denote its smallest eigenvalue.

\subsection{Preliminaries}
\label{prelim}
In this section we carryout some simple calculations yielding simple formulas for the gradient and Jacobian mappings. We begin by noting we can rewrite the gradient descent iterations in the form
\begin{align*}
\text{vect}\left(W_{\tau+1}\right)=\text{vect}\left(W_\tau\right)-\eta \text{vect}\left(\nabla \mathcal{L}\left(W_\tau\right)\right).
\end{align*}
Here,
\begin{align*}
\text{vect}\left(\nabla \mathcal{L}\left(W_\tau\right)\right)=\mathcal{J}^T(W_\tau)r\left(W_\tau\right),
\end{align*}
where
\begin{align*}
\mathcal{J}\left(W\right)=\frac{\partial }{\partial \text{vect}\left(W\right)}f(W)\quad\text{and}\quad 
\end{align*}
is the Jacobian mapping associated to the network and 
\begin{align*}
r\left(W\right)=\phi\left(V\phi\left(W z\right)\right)-y.
\end{align*}
is the misfit or residual vector. Note that
\begin{align*}
\frac{\partial}{\partial \text{vect}\left(W\right)}v^T\phi\left(Wz\right)=&
\begin{bmatrix}
v_1\phi'\left(w_1^Tz\right)x^T &
\ldots
&v_k\phi'\left(w_k^Tz\right)x^T
\end{bmatrix}\\
=&\left(v\odot\phi'\left(Wx\right)\right)^T\otimes x^T
\end{align*}
Thus
\begin{align*}
\mathcal{J}\left(W\right)=\left(V\text{diag}\left(\phi'(Wz)\right)\right)*\left(1z^T\right),
\end{align*}
This in turn yields
\begin{align}
\label{temp1}
\mathcal{J}\left(W\right)\mathcal{J}^T\left(W\right)=&\left(V\text{diag}\left(\phi'(W z)\right)\text{diag}\left(\phi'(Wz)\right)V^T\right)\nonumber
\ldots \\
& \odot\left(\twonorm{z}^211^T\right)\nonumber\\
=&\twonorm{z}^2V\text{diag}\left(\phi'(W z)\odot\phi'(W z)\right)V^T
\end{align}

\subsection{Lemmas for controlling the spectrum of the Jacobian and initial misfit}
\label{keyl}
In this section we state a few lemmas concerning the spectral properties of the Jacobian mapping, its perturbation and initial misfit of the model with the proofs deferred to Appendix \ref{appproof}.
\begin{lemma}[Minimum singular value of the Jacobian at initialization]\label{mineig} Let $V\in\R^{n\times d}$ and $W\in\R^{d\times k}$ be random matrices with i.i.d.~$\mathcal{N}(0,\nu^2)$ and $\mathcal{N}(0,1)$ entries and define the Jacobian mapping $\mathcal{J}\left(W\right)=\left(V\text{diag}\left(\phi'(Wz)\right)\right)*\left(1z^T\right)$. Then as long as $d\ge 3828n$,
\begin{align*}
\sigma_{\min}\left(\mathcal{J}(W)\right)\ge \frac{1}{2}\nu\sqrt{d}\twonorm{z}.
\end{align*}
holds with probability at least $1-2e^{-n}$.
\end{lemma}
\begin{lemma}[Perturbation lemma]\label{pert} Let $V\in\R^{n\times d}$ be a matrix with i.i.d.~$\mathcal{N}(0,\nu^2)$ entries, $W\in\R^{d\times k}$, and define the Jacobian mapping $\mathcal{J}\left(W\right)=\left(V\text{diag}\left(\phi'(Wz)\right)\right)*\left(1z^T\right)$. Also let $W_0$ be a matrix with i.i.d.~$\mathcal{N}(0,1)$ entries. Then, 
\begin{align*}
\opnorm{\mathcal{J}(W)-\mathcal{J}(W_0)}\le& \, \nu\twonorm{z}\Bigg( 2\sqrt{n}+
\ldots \\
&          
\sqrt{6\left(2dR\right)^{\frac{2}{3}}\log\bigg(\frac{d}{3\left(2dR\right)^{\frac{2}{3}}}\bigg)}\,\,\Bigg),
\end{align*}
holds for all $W\in\R^{d\times k}$ obeying $\opnorm{W-W_0}\le R$ with probability at least $1-e^{-n/2}-e^{-\frac{\left(2dR\right)^{\frac{2}{3}}}{6}}$.
\end{lemma}
\begin{lemma}[Spectral norm of the Jacobian]\label{spec} Let $V\in\R^{n\times d}$ be a matrix with i.i.d.~$\mathcal{N}(0,\nu^2)$ entries, $W\in\R^{d\times k}$, and define the Jacobian mapping $\mathcal{J}\left(W\right)=\left(V\text{diag}\left(\phi'(Wz)\right)\right)*\left(1z^T\right)$. Then,
\begin{align*}
\opnorm{\mathcal{J}(W)}\le\nu\left( \sqrt{d}+2\sqrt{n}\right)\twonorm{z},
\end{align*}
holds for all $W\in\R^{d\times k}$ with probability at least $1-e^{-n/2}$.
\end{lemma}
\begin{lemma}[Initial misfit]\label{init} Let $V\in\R^{n\times d}$ be a matrix with i.i.d.~$\mathcal{N}(0,\nu^2)$ entries with $\nu=\frac{1}{\sqrt{dn}}\frac{\twonorm{y}}{\twonorm{z}}$. Also let $W\in\R^{d\times k}$ be a matrix with i.i.d.~$\mathcal{N}(0,1)$ entries. Then
\begin{align*}
\twonorm{V\phi\left(Wz\right)-y}\le 3\twonorm{y},
\end{align*}
holds with probability at least $1-e^{-n/2}-e^{-d/2}$.
\end{lemma}

\subsection{Proof of Theorem \ref{mainthm}}
\label{pfmain}
Consider a nonlinear least-squares optimization problem of the form 
\begin{align*}
\underset{\theta\in\R^p}{\min}\text{ }\mathcal{L}(\theta):=\frac{1}{2}\twonorm{f(\theta)-y}^2,
\end{align*}
 with $f:\R^p\mapsto \R^n$ and $y\in\R^n$. Suppose the Jacobian mapping associated with $f$ obeys the following three assumptions.
 \begin{assumption} \label{ass1}Fix a point $\bteta_0$. We have that $\sigma_{\min}\left(\mathcal{J}(\bteta_0)\right)\geq 2\alpha$.
\end{assumption}
\begin{assumption}  \label{ass2}Let $\|\cdot\|$ denote a norm that is dominated by the Euclidean norm i.e.~$\|\theta\|\le \twonorm{\theta}$ holds for all $\theta\in
\R^p$. Fix a point $\bteta_0$ and a number $R>0$. For any $\bteta$ satisfying $\|\bteta-\bteta_0\|\leq R$, we have that $\|\mathcal{J}(\bteta_0)-\mathcal{J}(\bteta)\|\leq \alpha/3$.
\end{assumption}
\begin{assumption}  \label{ass3}For all $\bteta\in\R^p$, we have that $\|\mathcal{J}(\bteta)\|\leq \beta$.
\end{assumption}
Under these assumptions we can state the following theorem from \cite{oymak2019towards}.
\begin{theorem} [Non-smooth Overparameterized Optimization]\label{metathm}Given $\bteta_0\in\R^p$, suppose Assumptions \ref{ass1}, \ref{ass2}, and \ref{ass3} hold with 
\[
R=\frac{5\twonorm{y-f(\bteta_0)}}{\alpha}.
\]
Then, picking constant learning rate $\eta\leq \frac{1}{\beta^2}$, all gradient iterations obey the followings
\begin{align}
&\twonorm{y-f(\bteta_{\tau})}\leq (1-\frac{\eta\alpha^2}{4})^{\tau}\twonorm{y-f(\bteta_0)}\\
&\frac{\alpha}{5}\|\bteta_{\tau}-\bteta_0\|+\twonorm{y-f(\bteta_{\tau})}\leq \twonorm{y-f(\bteta_{0})}.
\end{align}
\end{theorem}
We shall apply this theorem to the case where the parameter is $W$ and the nonlinear mapping is given by $V\phi\left(Wz\right)$ and $\phi=ReLU$. All that is needed to be able to apply this theorem is check that the assumptions hold. Per the assumptions of the theorem we use 
\begin{align*}
\nu=\frac{1}{\sqrt{dn}}\frac{\twonorm{y}}{\twonorm{z}}.
\end{align*}
To this aim note that using Lemma \ref{mineig} Assumption \ref{ass1} holds with 
\begin{align*}
\alpha=\frac{1}{4}\nu\sqrt{d}\twonorm{z}=\frac{1}{4\sqrt{n}}\twonorm{y},
\end{align*}
with probability at least $1-2e^{-n}$. Furthermore, by Lemma \ref{spec} Assumption \ref{ass3} holds with 
\begin{align*}
\beta=&\,\frac{\twonorm{y}}{\sqrt{8n}}\sqrt{4n+d}
\\\ge\,&\frac{1}{2} \left(\sqrt{\frac{d}{4n}}+1\right)\twonorm{y}
\\=&\,\nu\left( \sqrt{d}+2\sqrt{n}\right)\twonorm{z}.
\end{align*}
with probability at least $1-e^{-n/2}$. All that remains for applying the theorem above is to verify Assumption \ref{ass2} holds with high probability
\begin{align*}
R=60\sqrt{n}= 15\frac{\twonorm{y}}{\alpha}\ge \frac{5}{\alpha}\twonorm{V\phi\left(Wz\right)-y}
\end{align*}
In the above we have used Lemma \ref{init} to conclude that $\twonorm{V\phi\left(Wz\right)-y}\le 3\twonorm{y}$ holds with probability at least $1-e^{-n/2}-e^{-d/2}$. Thus, using Lemma \ref{pert} all that remains is to show that 
\begin{align*}
\frac{1}{\sqrt{dn}}\twonorm{y}\left(2\sqrt{n}+\sqrt{6\left(2dR\right)^{\frac{2}{3}}\ln\left(\frac{d}{3\left(2dR\right)^{\frac{2}{3}}}\right)}\right)
\\\le&\, \frac{\alpha}{3}
\\=&\,\frac{\twonorm{y}}{12\sqrt{n}},
\end{align*}
holds with $R=60\sqrt{n}$ and with probability at least $1-e^{-n/2}-e^{-\frac{(120)^{\frac{2}{3}}}{6}d^{\frac{2}{3}}n^{\frac{1}{3}}}\ge 1-e^{-n/2}-e^{-4d^{\frac{2}{3}}n^{\frac{1}{3}}}$. The latter is equivalent to
\begin{align*}
2\sqrt{n}+\sqrt{6\left(120d\sqrt{n}\right)^{\frac{2}{3}}\ln\left(\frac{d}{3\left(120d\sqrt{n}\right)^{\frac{2}{3}}}\right)}\le \frac{\sqrt{d}}{12},
\end{align*}
which can be rewritten in the form
\begin{align*}
2\sqrt{\frac{n}{d}}+\sqrt{6\left(120\right)^{\frac{2}{3}} \sqrt[3]{\frac{n}{d}}\ln\left(\frac{1}{3(120)^{\frac{2}{3}}\sqrt[3]{\frac{n}{d}}}\right)}\le \frac{1}{12},
\end{align*}
which holds as long as $d\ge 4.3\times 10^{15} n$. Thus with $d\ge Cn$ then Assumptions \ref{ass1}, \ref{ass2}, and \ref{ass3} holds with probability at least $1-5e^{-n/2}-e^{-d/2}-e^{-4d^{\frac{2}{3}}n^{\frac{1}{3}}}$. Thus, Theorem \ref{metathm} holds with high probability. Applying Theorem \ref{metathm} completes the proof.

\section{Proof of Lemmas for the Spectral Properties of the Jacobian}
\label{appproof}
\subsection{Proof of Lemma \ref{mineig}}
We prove the result for $\nu=1$, the general result follows from a simple re-scaling. Define the vectors
\begin{align*}
a_\ell= V_\ell\phi'\left(\langle w_\ell,z\rangle\right)\in\R^n,
\end{align*}
with $V_\ell$ the $\ell$th column of $V$. Using \eqref{temp1} we have
\begin{align}
\label{myinter}
\mathcal{J}\left(W\right)\mathcal{J}^T\left(W\right)=&\twonorm{z}^2V\text{diag}\left(\phi'(Wz)\odot\phi'(Wz)\right)V^T,\nonumber\\
=&\twonorm{z}^2\left(\sum_{\ell=1}^d a_\ell a_\ell^T\right),\nonumber\\
=&d\twonorm{z}^2\left(\frac{1}{d}\sum_{\ell=1}^d a_\ell a_\ell^T\right).
\end{align}
To bound the minimum eigenvalue we state a result from \cite{oliveira2013lower}.
\begin{theorem}\label{mineigthm} Assume $A_1,\ldots,A_d\in\R^{n\times n}$ are i.i.d.~random positive semidefinite matrices whose coordinates have bounded second moments. Define $\Sigma:=\E[A_1]$ (this is an entry-wise expectation) and 
\begin{align*}
\widehat{\Sigma}_d=\frac{1}{d}\sum_{\ell=1}^dA_\ell.
\end{align*}
Let $h\in(1,+\infty)$ be such that $\sqrt{\E\big[\left(u^TA_1u\right)^2\big]}\le h u^T\Sigma u$ for all $u\in\R^n$. Then for any $\delta\in(0,1)$ we have
\begin{align*}
\mathbb{P}\Bigg\{\forall u\in\R^n: u^T\widehat{\Sigma}_ku\ge \left(1-7h\sqrt{\frac{n+2\ln(2/\delta)}{d}}\right)u^T\Sigma u\Bigg\}\\\ge 1-\delta
\end{align*}
\end{theorem}
We shall apply this theorem with $A_\ell:=a_\ell a_\ell^T$. To do this we need to calculate the various parameters in the theorem. We begin with $\Sigma$ and note that for ReLU we have
\begin{align*}
\Sigma:=&\E[A_1]\\
=&\E\big[a_1a_1^T\big]\\
=&\E_{w\sim\mathcal{N}(0,I_k)}\big[\left(\phi'(\langle w,z\rangle)\right)^2\big]\E_{v\sim\mathcal{N}(0,I_n)}[vv^T]\\
=&\E_{w\sim\mathcal{N}(0,I_k)}\big[\left(\phi'(w^Tz)\right)^2\big]I_n\\
=&\E_{w\sim\mathcal{N}(0,I_k)}\big[\mathbb{I}_{\{w^Tz\ge 0\}}\big]I_n\\
=&\frac{1}{2}I_n.
\end{align*}
To calculate $h$ we have
\begin{align*}
\sqrt{\E\big[\left(u^TA_1u\right)^2\big]}\le&\sqrt{\E\Big[\left(a_1^Tu\right)^4\Big]}\\
\le& \sqrt{\E_{w\sim\mathcal{N}(0,I_k)}\big[\mathbb{I}_{\{w^Tz\ge 0\}}\big]\cdot\E_{v\sim\mathcal{N}(0,I_n)}\Big[\left(v^Tu\right)^4\Big]}\\
\le&\sqrt{\frac{3}{2}\twonorm{u}^4}\\
\le&\frac{\sqrt{3}}{\sqrt{2}}\twonorm{u}^2\\
=&\sqrt{6}u^T\left(\frac{1}{2}I_n\right)u\\
=&\sqrt{6}\cdot u^T\Sigma u.
\end{align*}
Thus we can take $h=\sqrt{6}$. Therefore, using Theorem \ref{mineigthm} with $\delta=2 e^{-n}$ we can conclude that
\begin{align*}
\lambda_{\min}\left(\frac{1}{d}\sum_{\ell=1}^d a_\ell a_\ell^T\right)\ge \frac{1}{4} 
\end{align*}
holds with probability at least $1-2e^{-n}$ as long as
\begin{align*}
d\ge 3528\cdot n.
\end{align*}
Plugging this into \eqref{myinter} we conclude that with probability at least $1-2e^{-n}$
\begin{align*}
\sigma_{\min}\left(\mathcal{J}(W)\right)\ge \frac{1}{2}\sqrt{d}\twonorm{z}.
\end{align*}
\subsection{Proof of Lemma \ref{pert}}
We prove the result for $\nu=1$, the general result follows from a simple rescaling. Based on \eqref{temp1} we have
\begin{align*}
&\left(\mathcal{J}(W)-\mathcal{J}(W_0)\right)\left(\mathcal{J}(W)-\mathcal{J}(W_0)\right)^T \ldots
\\&=\twonorm{z}^2V\text{diag}\big(\left(\phi'(Wz)-\phi'(W_0z)\right)\odot \ldots 
\\& \;\;\;\; \left(\phi'(Wz)-\phi'(W_0z)\right)\big)V^T.
\end{align*}
Thus
\begin{align}
\label{jpert}
\opnorm{\mathcal{J}(W)-\mathcal{J}(W_0)}\le& \twonorm{z}\opnorm{V\text{diag}\left(\phi'(W z)-\phi'(W_0z)\right)}\\
=&\twonorm{z}\opnorm{V\text{diag}\left(\mathbb{I}_{\{Wz\ge 0\}}-\mathbb{I}_{\{W_0z\ge 0\}}\right)}\nonumber\\
\le &\twonorm{z}\opnorm{V_{\mathcal{S}(W)}},
\end{align}
where $\mathcal{S}(W)\subset \{1,2,\ldots,d\}$ is the set of indices where $Wz$ and $W_0z$ have different signs i.e.~$\mathcal{S}(W):=\{\ell: \sgn{e_\ell^TWz}\neq \sgn{e_\ell^TW_0z}\}$ and $V_{\mathcal{S}(W)}$ is a submatrix $V$ obtained by picking the columns corresponding to $\mathcal{S}(W)$.

To continue further note that by Gordon's lemma we have
\begin{align*}
\sup_{|\mathcal{S}|\le s}\opnorm{V_{\mathcal{S}}}\le \sqrt{n}+\sqrt{2s\log(d/s)}+t,
\end{align*}
with probability at least $1-e^{-t^2/2}$. In particular using $t=\sqrt{n}$ we conclude that
\begin{align}
\label{gordon}
\sup_{|\mathcal{S}|\le s}\opnorm{V_{\mathcal{S}}}\le 2\sqrt{n}+\sqrt{2s\log(d/s)},
\end{align}
with probability at least $1-e^{-n/2}$. To continue further we state a lemma controlling the size of $|\mathcal{S}(W)|$ based on the size of the radius $R$.
\begin{lemma}[sign changes in local neighborhood]\label{signch} Let $W_0\in\R^{d\times k}$ be a matrix with i.i.d.~$\mathcal{N}(0,1)$ entries. Also for a matrix $W\in\R^{d\times k}$ define $\mathcal{S}(W):=\{\ell: \sgn{e_\ell^T Wz}\neq \sgn{e_\ell^T W_0z}\}$. Then for any $W\in\R^{d\times k}$ obeying $\opnorm{W-W_0}\le R$
\begin{align*}
|\mathcal{S}(W)|\le 2\lceil \left(2dR\right)^{\frac{2}{3}} \rceil
\end{align*}
holds with probability at least $1-e^{-\frac{\left(2dR\right)^{\frac{2}{3}}}{6}}$.
\end{lemma}
Combining \eqref{jpert} together with \eqref{gordon} (using $s=3\left(2dR\right)^{\frac{2}{3}}$) and Lemma \ref{signch} we conclude that 
\begin{align*}
&\opnorm{\mathcal{J}(W)-\mathcal{J}(W_0)} \ldots
\\&\le \twonorm{z}\left(2\sqrt{n}+\sqrt{6\left(2dR\right)^{\frac{2}{3}}\log\left(\frac{d}{3\left(2dR\right)^{\frac{2}{3}}}\right)}\right)
\end{align*}
holds with probability at least $1-e^{-n/2}-e^{-\frac{\left(2dR\right)^{\frac{2}{3}}}{6}}$.
\subsection{Proof of Lemma \ref{signch}}
To prove this result we utilize two lemmas from \cite{oymak2019towards}. In these lemmas we use $|v|_{m-}$ to denote the $m$th smallest entry of $v$ after sorting its entries in terms of absolute value.
\begin{lemma}[Lemma C.2 in~\cite{oymak2019towards}] Given an integer $m$, suppose
\begin{align*}
\opnorm{W-W_0}\le \sqrt{m}\frac{|W_0z|_{m-}}{\twonorm{z}},
\end{align*}
then 
\begin{align*}
|\mathcal{S}(W)|\le 2m.
\end{align*}
\end{lemma}
\begin{lemma}[Lemma C.3 in~\cite{oymak2019towards}] Let $z\in\R^k$. Also let $W_0\in\R^{d\times k}$ be a matrix with i.i.d.~$\mathcal{N}(0,1)$ entries. Then, with probability at least $1-e^{-\frac{m}{6}}$,
\begin{align*}
\frac{|W_0z|_{m-}}{\twonorm{z}}\ge \frac{m}{2d}.
\end{align*}
\end{lemma}
Combining the latter two lemmas with $m=\lceil \left(2dR\right)^{\frac{2}{3}} \rceil$ we conclude that when 
\begin{align*}
\opnorm{W-W_0}\le& R\\
\le& \frac{m^{\frac{3}{2}}}{2d}\\
\le& \sqrt{m}\frac{m}{2d}\\
\le& \sqrt{m}\frac{|W_0z|_{m-}}{\twonorm{z}},
\end{align*}
then with probability at least $1-e^{-\frac{\left(2dR\right)^{\frac{2}{3}}}{6}}$ we have 
\begin{align*}
|\mathcal{S}(W)|\le 2m\le 2\lceil \left(2dR\right)^{\frac{2}{3}} \rceil.
\end{align*}
\subsection{Proof of Lemma \ref{spec}}
We prove the result for $\nu=1$, the general result follows from a simple rescaling. Using \eqref{temp1} we have
\begin{align*}
\mathcal{J}\left(W\right)\mathcal{J}^T\left(W\right)=&\twonorm{z}^2V\text{diag}\left(\phi'(Wz)\odot\phi'(Wz)\right)V^T
\end{align*}
Thus
\begin{align*}
\opnorm{\mathcal{J}\left(W\right)}\le& \twonorm{z}\opnorm{V\text{diag}\left(\phi'(Wz)\right)}\\
\le& \twonorm{z}\opnorm{V}
\end{align*}
The proof is complete by using standard concentration results for the spectral norm of a Gaussian matrix that allow us to conclude that
\begin{align*}
\opnorm{V}\le \sqrt{d}+2\sqrt{n},
\end{align*}
holds with probability at least $1-e^{-n/2}$.
\subsection{Proof of Lemma \ref{init}}
By the triangular inequality we have
\begin{align}
\label{inter2}
\twonorm{V\phi\left(Wz\right)-y}\le \twonorm{V\phi\left(Wz\right)}+\twonorm{y}
\end{align} 
To continue further let us consider one entry of $V\phi\left(Wz\right)$ and note that it has the same distribution as 
\begin{align*}
V\phi\left(Wz\right)\sim \nu\twonorm{\phi(Wz)}g,
\end{align*}
where $g\in\R^d$ is random Gaussian vectors with distribution $g\sim\mathcal{N}(0,I_d)$. Thus
\begin{align}
\label{inter}
\twonorm{V\phi\left(Wz\right)}\sim \nu\twonorm{\phi(Wz)}\twonorm{g}
\le& \sqrt{2n}\nu\twonorm{\phi(Wz)}
\\\le& \sqrt{2n}\nu\twonorm{Wz},
\end{align}
with probability at least $1-e^{-n/2}$. Furthermore, note that
\begin{align*}
Wz\sim\twonorm{z}\widetilde{g},
\end{align*}
where $\widetilde{g}\in\R^d$ is random Gaussian vectors with distribution $\widetilde{g}\sim\mathcal{N}(0,I_d)$. Combining the latter with \eqref{inter} we conclude that 
\begin{align*}
\twonorm{V\phi\left(Wz\right)}\le 2\sqrt{nd}\nu\twonorm{z}=2\twonorm{y},
\end{align*}
holds with probability at least $1-e^{-n/2}-e^{-d/2}$. Combining the latter with \eqref{inter2} we conclude that
\begin{align*}
\twonorm{V\phi\left(Wz\right)-y}\le 3\twonorm{y},
\end{align*}
holds with probability at least $1-e^{-n/2}-e^{-d/2}$.
\clearpage

\newsavebox\myboxfour
\savebox{\myboxfour}{\includegraphics[width=0.46\textwidth]{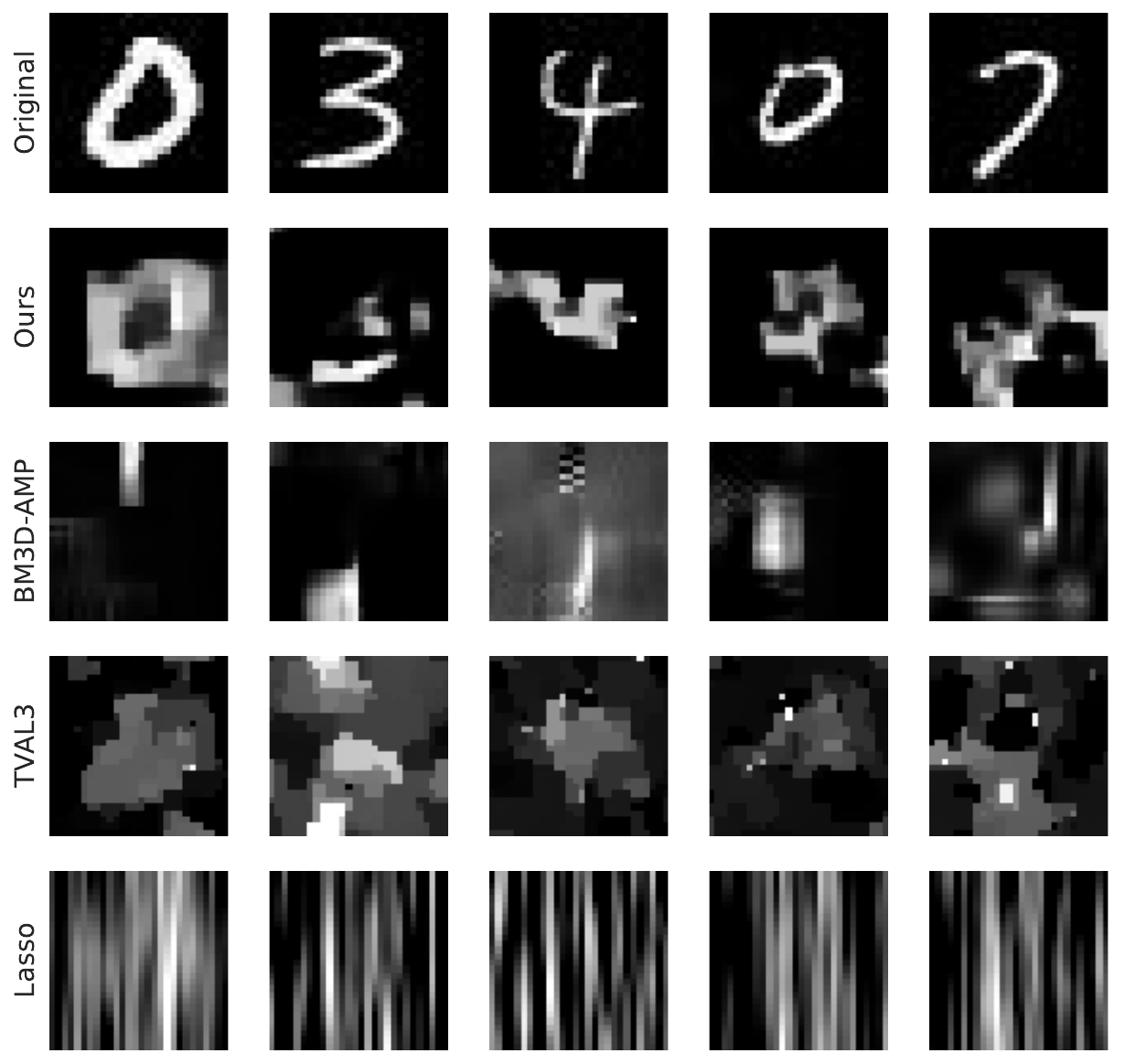}}
\begin{figure*}[b!]
    \begin{subfigure}[t]{0.46\textwidth}
        \usebox{\myboxfour}
        \vspace*{-3mm}
        \caption{25 measurements}
    \end{subfigure}\hfill%
    \begin{subfigure}[t]{0.46\textwidth}
        \vbox to \ht\myboxfour{%
            \vfill \includegraphics[width=\textwidth]{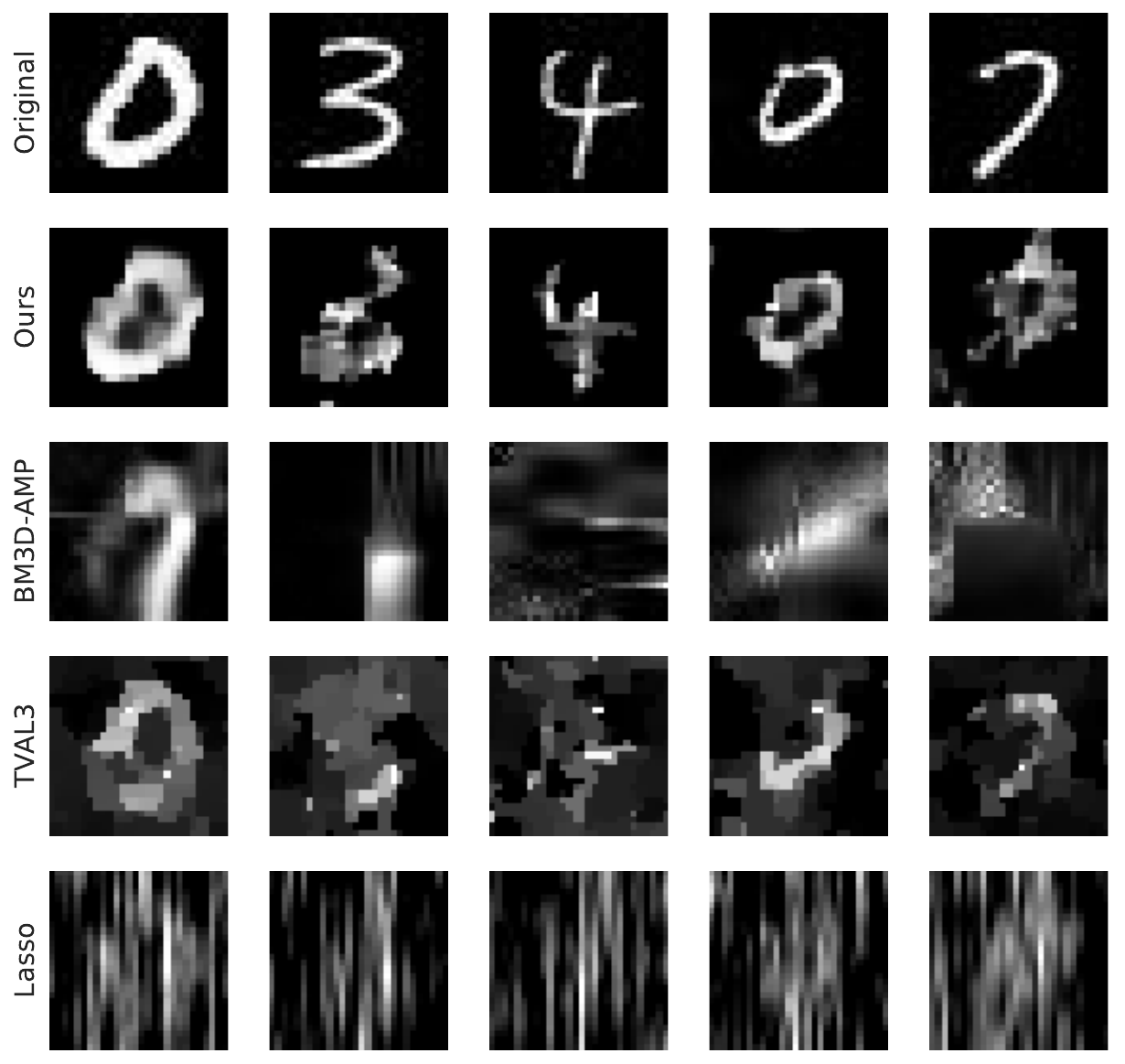}
        }
        \vspace*{2mm}
        \caption{50 measurements}
    \end{subfigure}
    \caption{Reconstruction results on MNIST for m = 25, 50 measurements respectively (of n = 784 pixels). From top to bottom row: original image, reconstructions by our algorithm, then reconstructions by baselines BM3D-AMP, TVAL3, and Lasso.} 
\end{figure*}

\newsavebox\myboxfive
\savebox{\myboxfive}
{\includegraphics[width=0.46\textwidth]{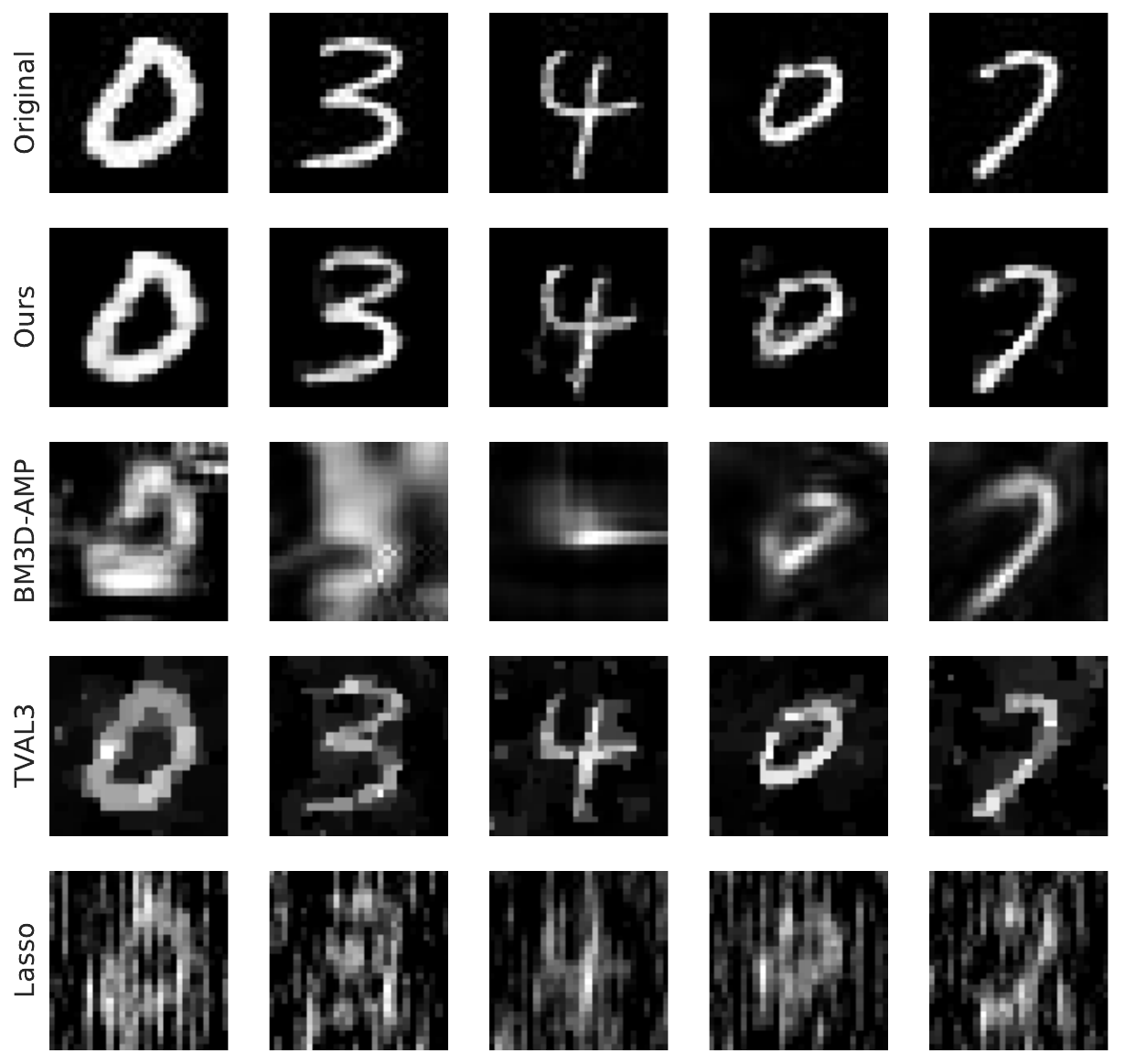}}
\begin{figure*}
    \begin{subfigure}[t]{0.46\textwidth}
        \usebox{\myboxfive}
        \vspace*{-3mm}
        \caption{100 measurements}
    \end{subfigure}\hfill%
    \begin{subfigure}[t]{0.46\textwidth}
        \vbox to \ht\myboxfive{%
            \vfill \includegraphics[width=\textwidth]{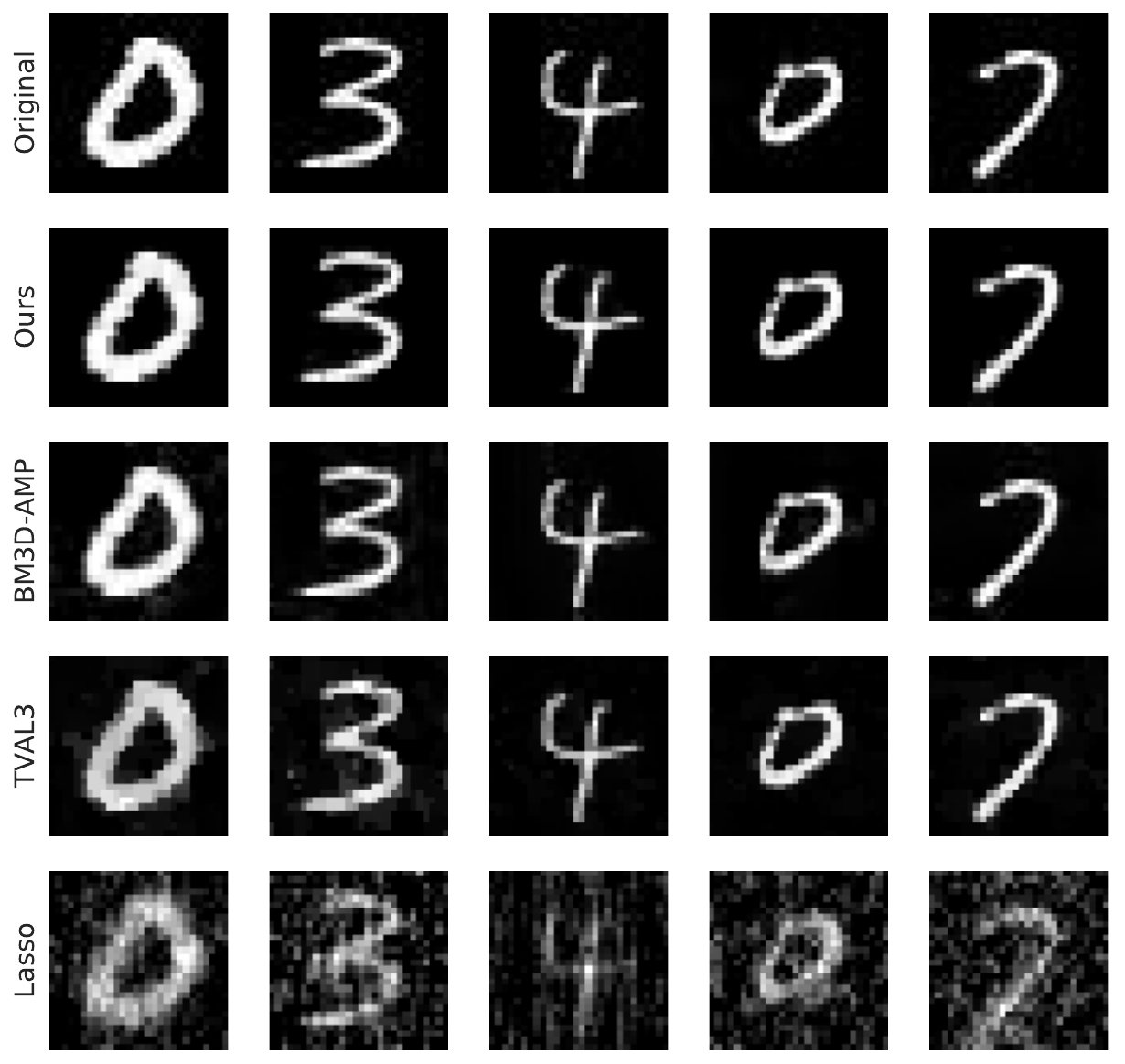}
        }
        \vspace*{2mm}
        \caption{200 measurements}
    \end{subfigure}
    \caption{Reconstruction results on MNIST for m = 100, 200 measurements respectively (of n = 784 pixels). From top to bottom row: original image, reconstructions by our algorithm, then reconstructions by baselines BM3D-AMP, TVAL3, and Lasso.} 
\label{fig:recons_x_app1}
\end{figure*}
\newsavebox\myboxsix
\savebox{\myboxsix}{\includegraphics[width=0.46\textwidth]{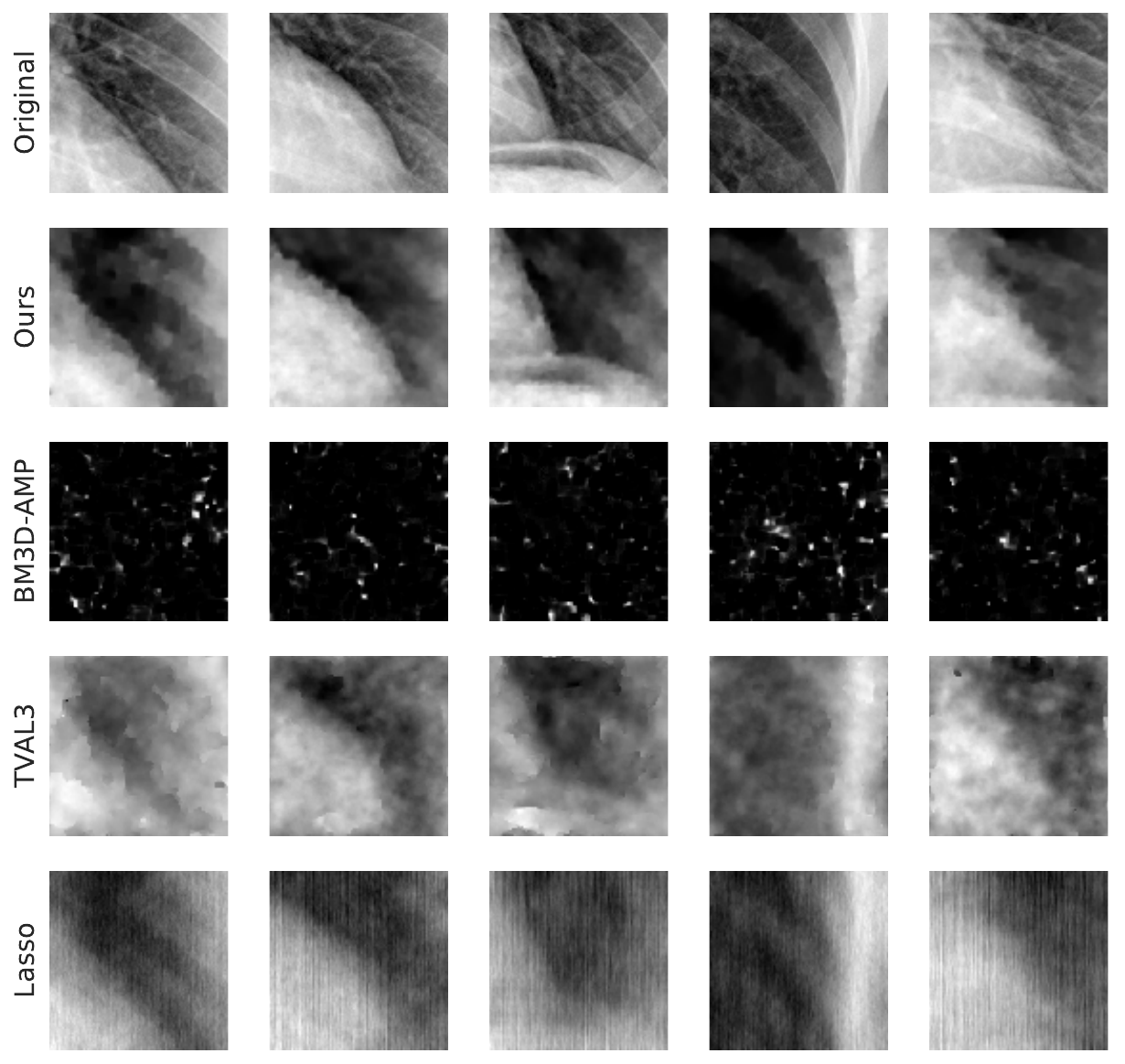}}
\begin{figure*}
    \begin{subfigure}[t]{0.46\textwidth}
        \usebox{\myboxsix}
        \vspace*{-3mm}
        \caption{500 measurements}
    \end{subfigure}\hfill%
    \begin{subfigure}[t]{0.46\textwidth}
        \vbox to \ht\myboxsix{%
            \vfill \includegraphics[width=\textwidth]{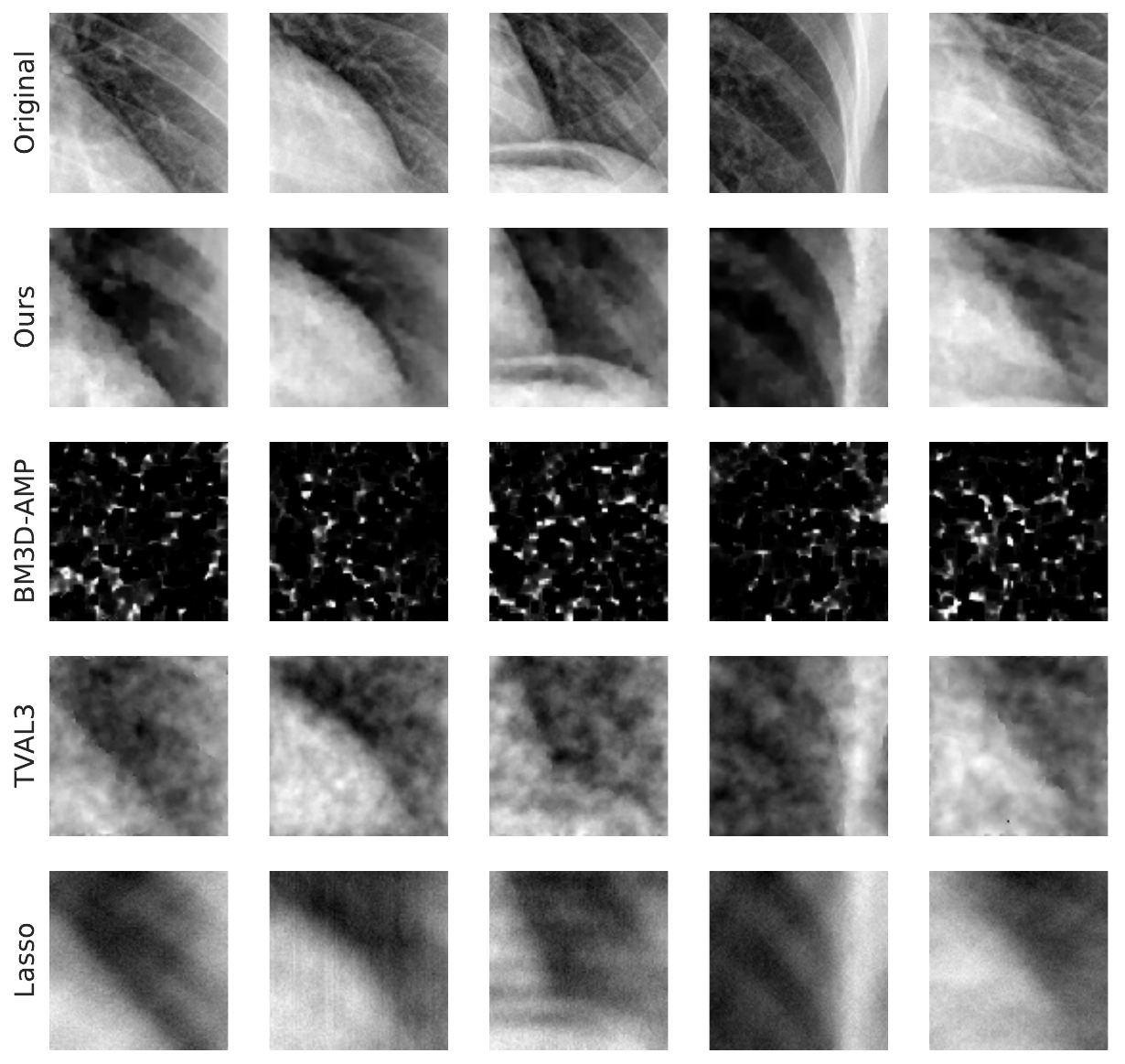}
        }
        \vspace*{2mm}
        \caption{1000 measurements}
    \end{subfigure}
    \caption{Reconstruction results on x-ray images for m = 500, 1000 measurements respectively (of n = 65536 pixels). From top to bottom row: original image, reconstructions by our algorithm, then reconstructions by baselines BM3D-AMP, TVAL3, and Lasso.} 
\end{figure*}

\newsavebox\myboxseven
\savebox{\myboxseven}
{\includegraphics[width=0.46\textwidth]{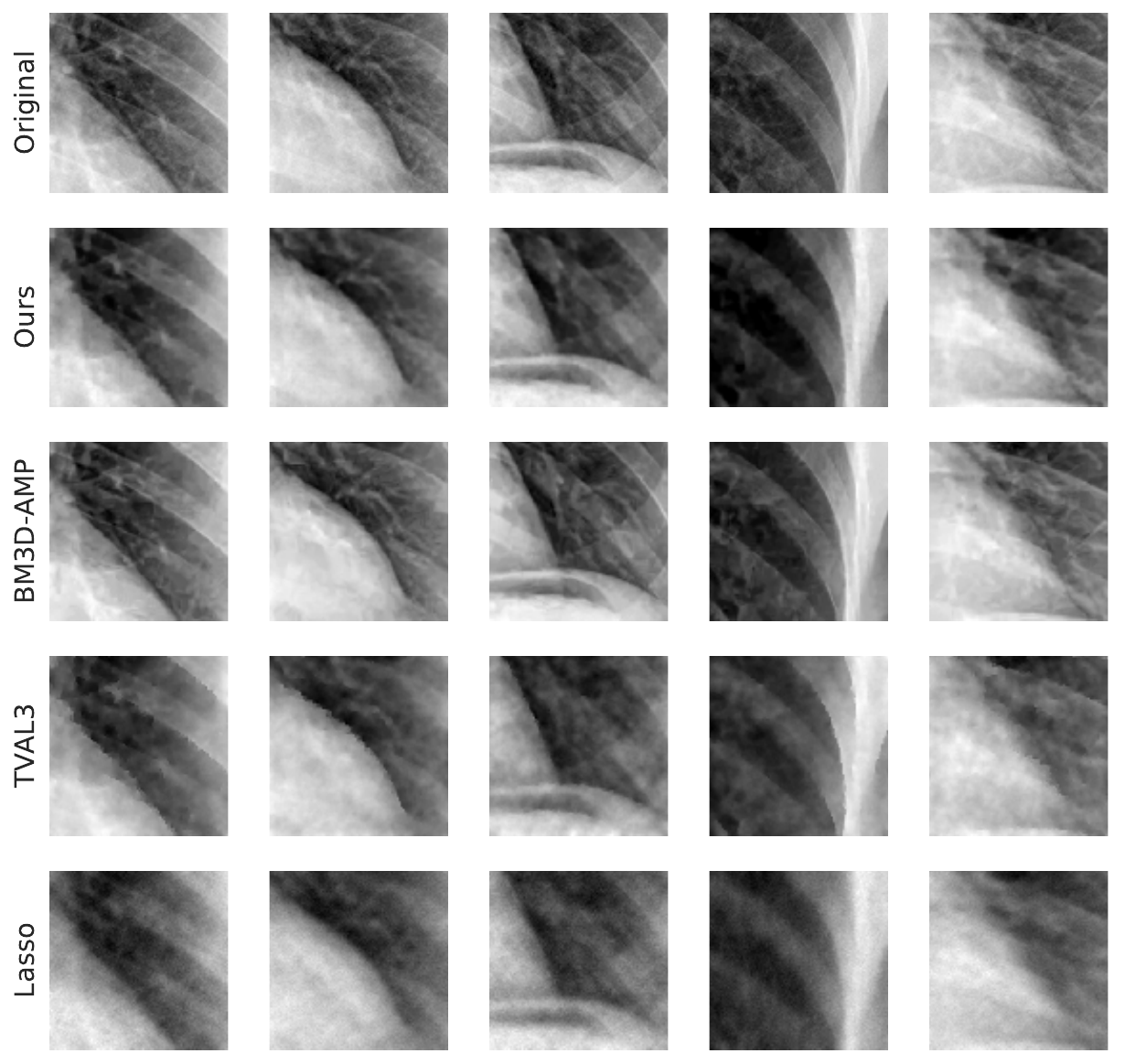}}
\begin{figure*}
    \begin{subfigure}[t]{0.46\textwidth}
        \usebox{\myboxseven}
        \vspace*{-3mm}
        \caption{4000 measurements}
    \end{subfigure}\hfill%
    \begin{subfigure}[t]{0.46\textwidth}
        \vbox to \ht\myboxseven{%
            \vfill \includegraphics[width=\textwidth]{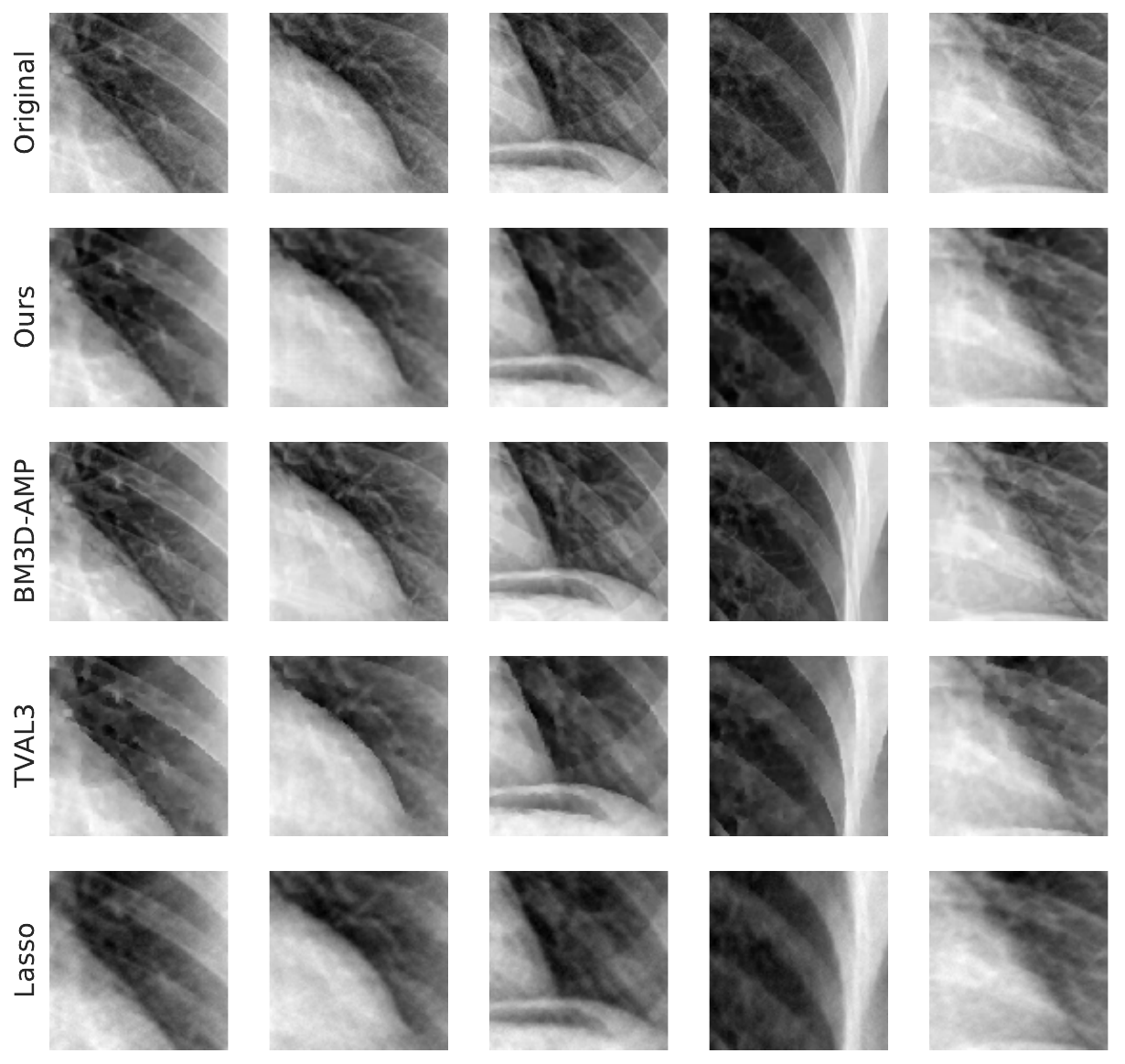}
        }
        \vspace*{2mm}
        \caption{8000 measurements}
    \end{subfigure}
    \caption{Reconstruction results on x-ray images for m = 4000, 8000 measurements respectively (of n = 65536 pixels). From top to bottom row: original image, reconstructions by our algorithm, then reconstructions by baselines BM3D-AMP, TVAL3, and Lasso.} 
\end{figure*}
\newsavebox\myboxeight
\savebox{\myboxeight}
{\includegraphics[width=0.46\textwidth]{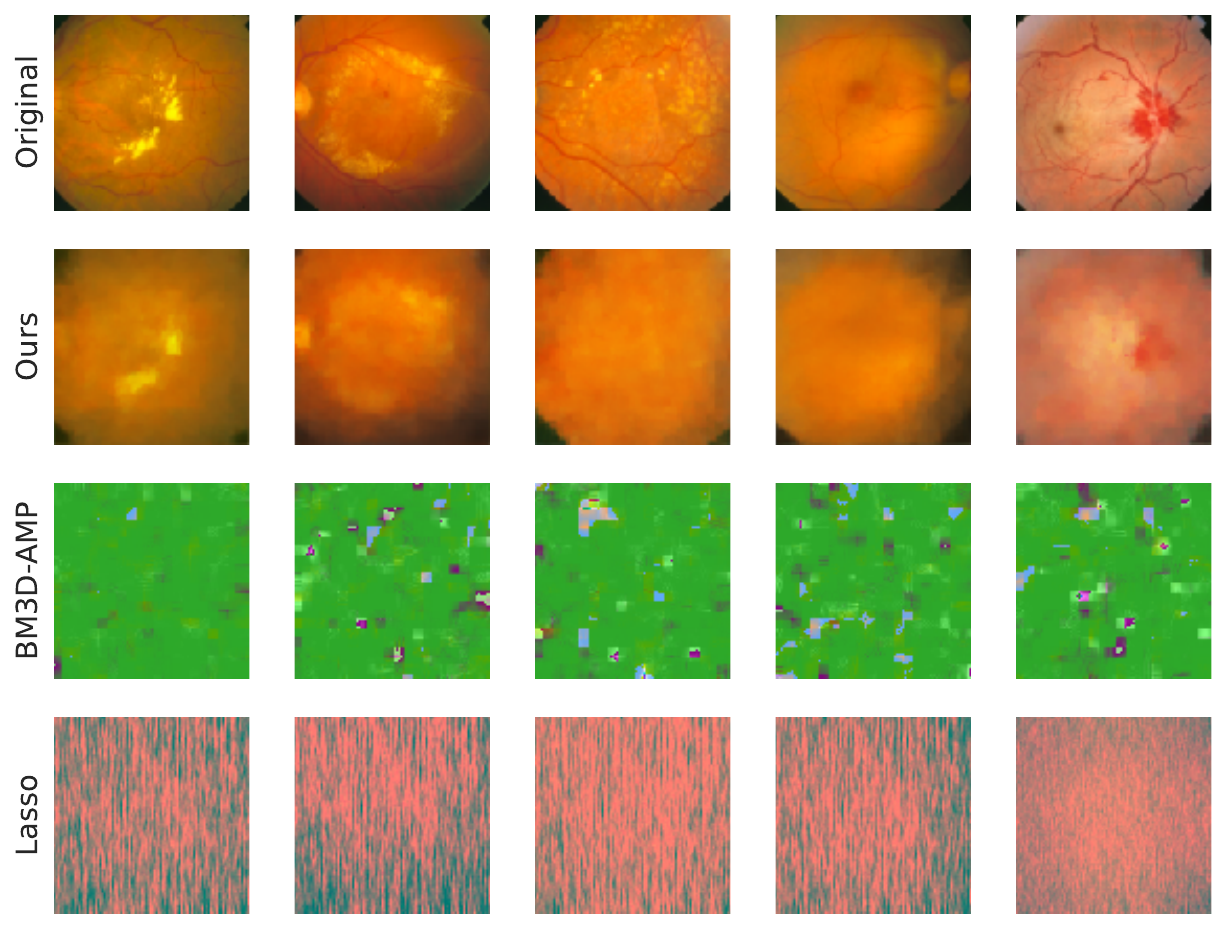}}
\begin{figure*}[b!]
    \begin{subfigure}[t]{0.46\textwidth}
        \usebox{\myboxeight}
        \vspace*{-3mm}
        \caption{500 measurements}
    \end{subfigure}\hfill%
    \begin{subfigure}[t]{0.46\textwidth}
        \vbox to \ht\myboxeight{%
            \vfill \includegraphics[width=\textwidth]{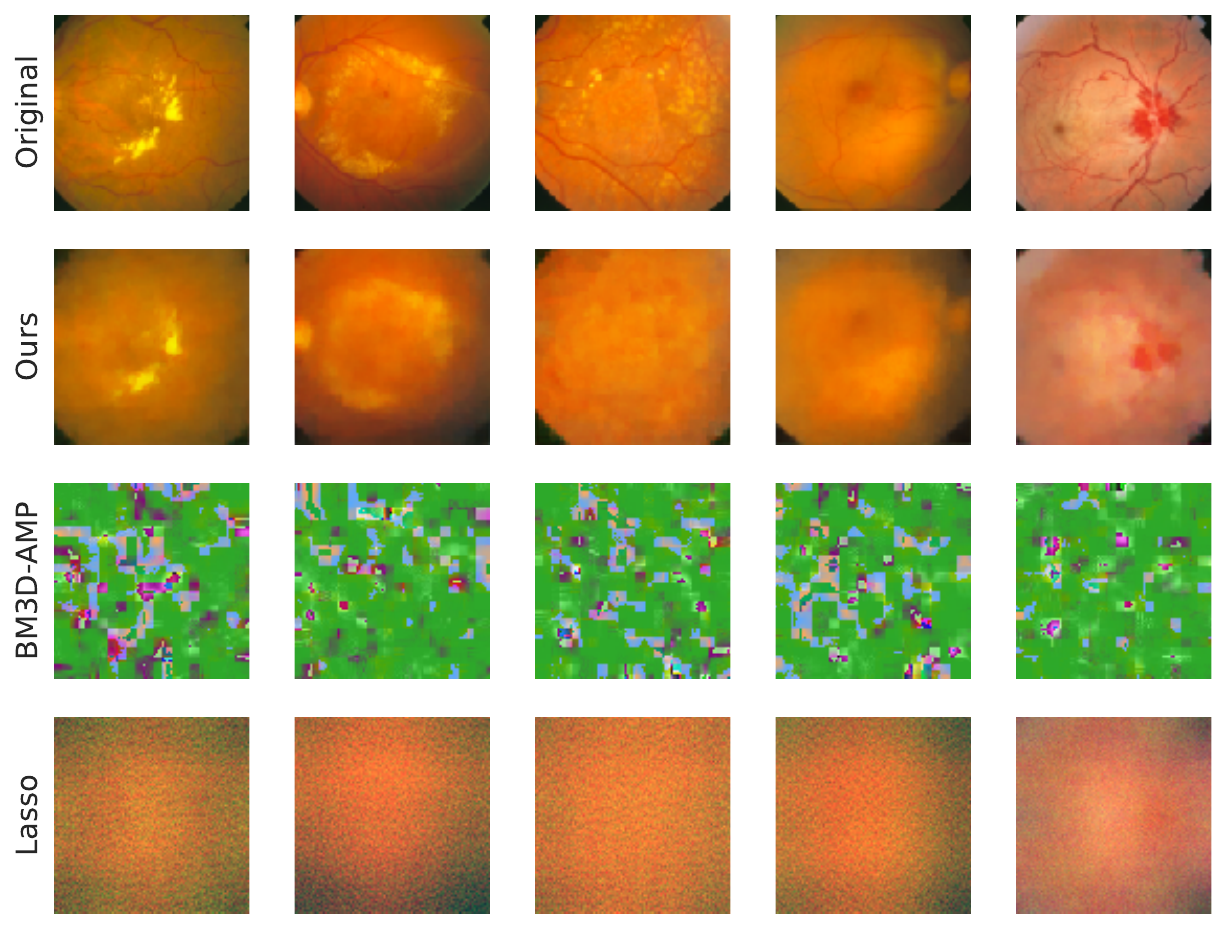}
        }
        \vspace*{2mm}
        \caption{1000 measurements}
    \end{subfigure}
    \caption{Reconstruction results on retinopathy images for m = 500, 1000 measurements respectively (of n = 49152 pixels). From top to bottom row: original image, reconstructions by our algorithm, then reconstructions by baselines BM3D-AMP and Lasso.} 
\end{figure*}
\begin{figure*}[t]
\vskip 0.2in
\begin{center}
\centerline{\includegraphics[width=0.8\textwidth]{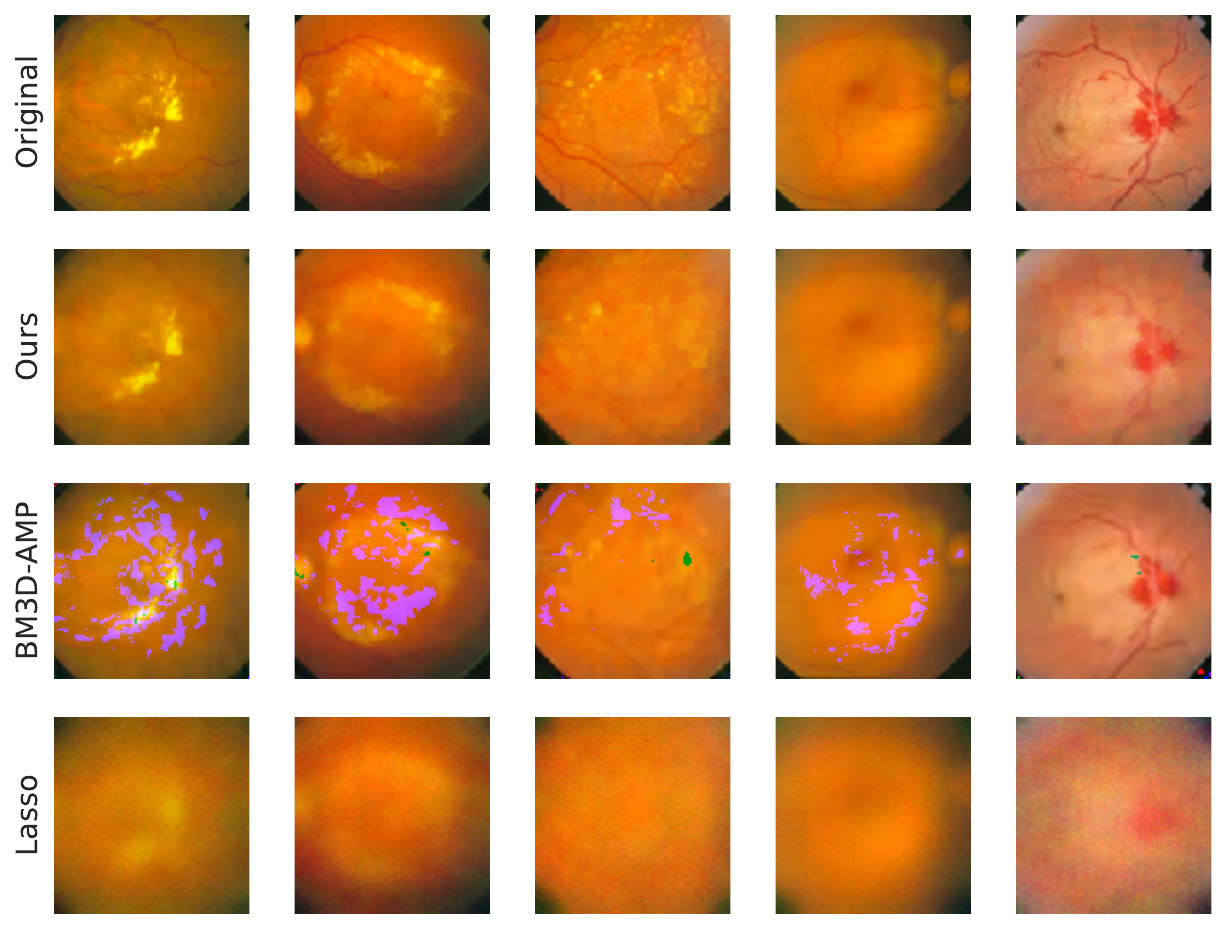}}
\caption{Reconstruction results on retinopathy images for m = 2000 measurements (of n = 49152 pixels). From top to bottom row: original image, reconstructions by our algorithm, then reconstructions by baselines BM3D-AMP and Lasso. In this case the number of measurements is much smaller than the number of pixels (roughly 4\% ratio), for which BM3D-AMP fails to converge, as demonstrated by erroneous green and purple pixels. We recommend viewing in color.}
\label{fig:recons_r_standalone}
\end{center}
\vskip -0.2in
\end{figure*}
\begin{figure*}[ht]
\vskip 0.2in
\begin{center}
\centerline{\includegraphics[width=\columnwidth]{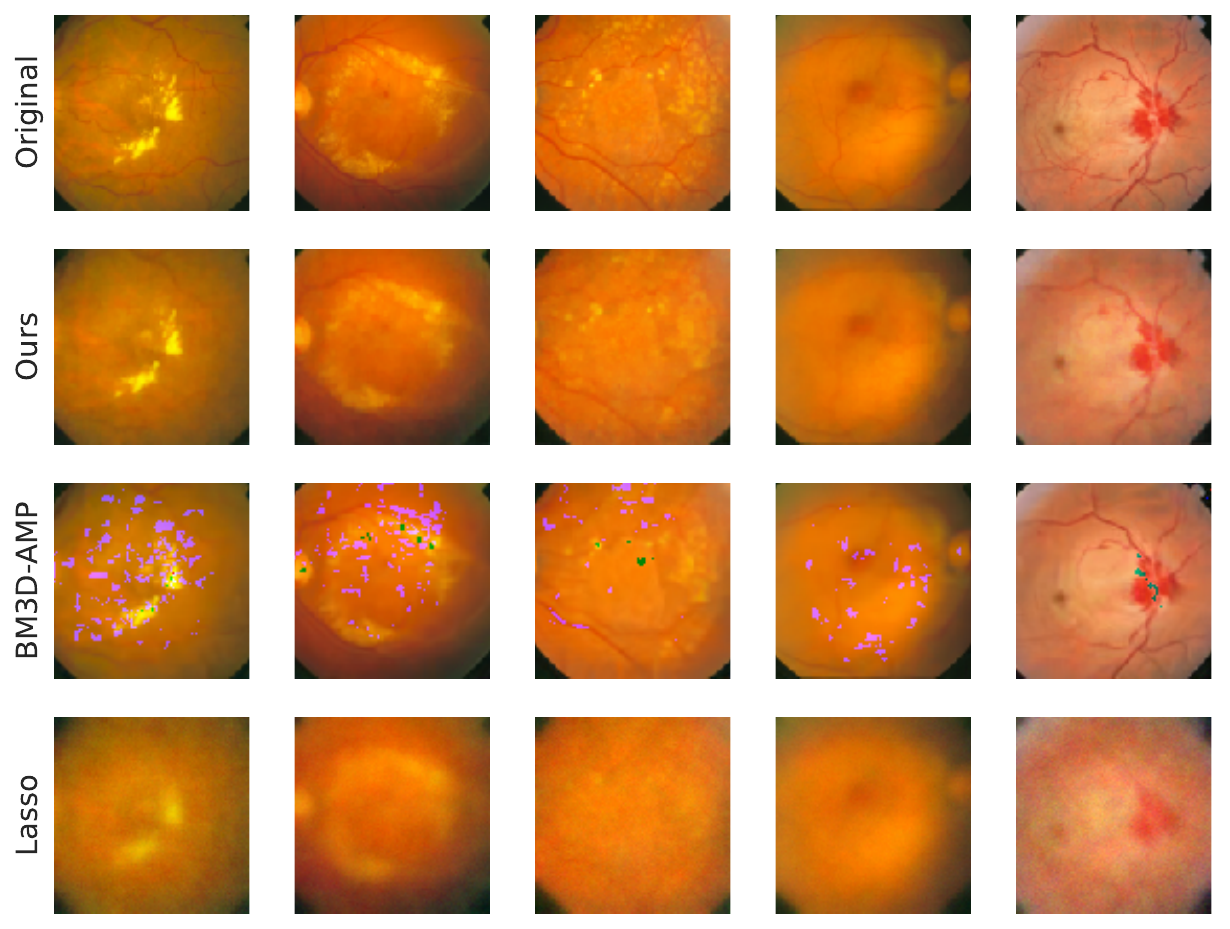}}
\caption{Reconstruction results on retinopathy images for m = 4000 (of n = 49152 pixels). From top to bottom row: original image, reconstructions by our algorithm, then reconstructions by baselines BM3D-AMP and Lasso.}
\end{center}
\vskip -0.2in
\end{figure*}

\begin{figure*}[ht]
\vskip 0.2in
\begin{center}
\centerline{\includegraphics[width=\columnwidth]{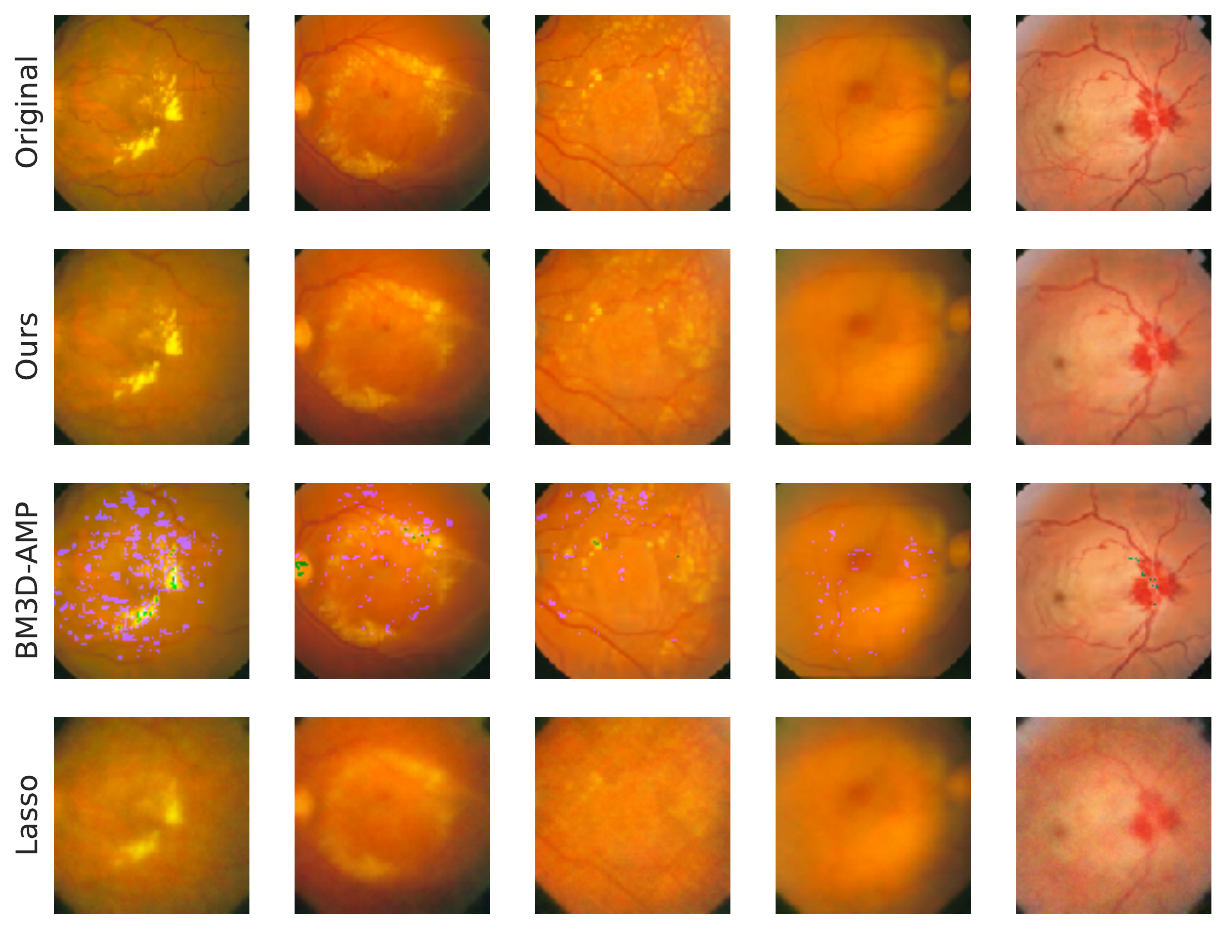}}
\caption{Reconstruction results on retinopathy images for m = 8000 (of n = 49152 pixels). From top to bottom row: original image, reconstructions by our algorithm, then reconstructions by baselines BM3D-AMP and Lasso.}
\end{center}
\vskip -0.2in
\end{figure*}
\clearpage
\begin{figure*}[t]
\vskip 0.2in
\begin{center}
\centerline{\includegraphics[width=\columnwidth]{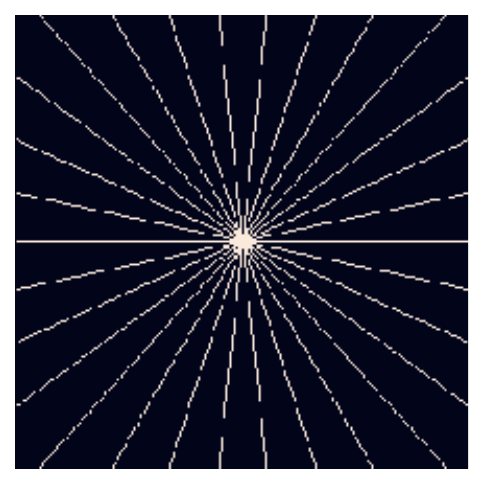}}
\caption{A radial sampling pattern of coefficients $\Omega$ in the Fourier domain. The measurements are obtained by sampling Fourier coefficients along these radial lines.}
\label{fig: fourier sampling pattern}
\end{center}
\vskip -0.2in
\end{figure*}

\begin{figure*}
\vskip 0.2in
\begin{center}
\centerline{\includegraphics[width=\columnwidth]
{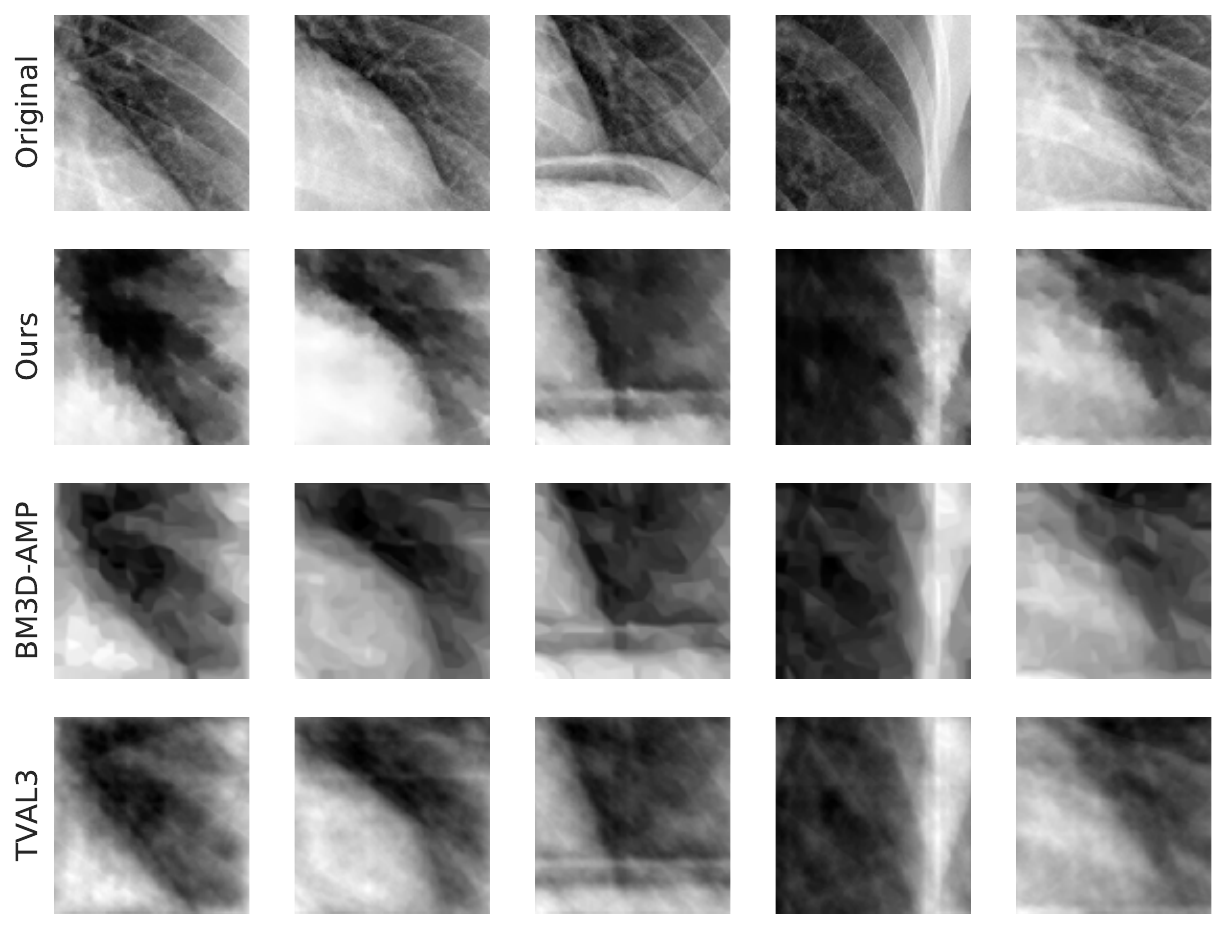}}
\caption{Reconstruction results on x-ray images for m = 1260 Fourier coefficients (of n = 65536 pixels). From top to bottom row: original image, reconstructions by our algorithm, then reconstructions by baselines BM3D-AMP and TVAL3.}
\label{fig: fourier reconstruction 10}
\end{center}
\vskip -0.2in
\end{figure*}
}

\end{document}